\documentclass[final]{cvpr}
\makeatletter
\@namedef{ver@everyshi.sty}{}
\makeatother
\usepackage{times}
\usepackage{epsfig}
\usepackage{fnpct}
\usepackage{graphicx}
\usepackage{amsmath, bm}
\usepackage{amssymb}
\usepackage{csvsimple}
\usepackage{verbatim}
\usepackage[pagebackref=true,breaklinks=true,colorlinks,bookmarks=false]{hyperref}

\def\cvprPaperID{****} 
\def\confYear{CVPR 2021}

\begin{document}

\title{Do All MobileNets Quantize Poorly? Gaining Insights into the Effect of Quantization on Depthwise Separable Convolutional Networks Through the Eyes of Multi-scale Distributional Dynamics}

\author{Stone Yun$^{1,2}$ and Alexander Wong$^{1,2}$\\
$^1$Vision and Image Processing Research Group, University of Waterloo\\
$^2$Waterloo Artificial Intelligence Institute, Waterloo, Canada\\
{\tt\small {\{s22yun, a28wong\}}@uwaterloo.ca}
}

\maketitle

\begin{abstract}
    As the ``Mobile AI" revolution continues to grow, so does the need to understand the behaviour of edge-deployed deep neural networks.  In particular, MobileNets~\cite{mbnetv1, mbnetv2} are the go-to family of deep convolutional neural networks (CNN) for mobile. However, they often have significant accuracy degradation under post-training quantization. While studies have introduced quantization-aware training and other methods to tackle this challenge, there is limited understanding into why MobileNets (and potentially depthwise-separable CNNs (DWSCNN) in general) quantize so poorly compared to other CNN architectures.  Motivated to gain deeper insights into this phenomenon, we take a different strategy and study the multi-scale distributional dynamics of MobileNet-V1, a set of smaller DWSCNNs, and regular CNNs.  Specifically, we investigate the impact of quantization on the weight and activation distributional dynamics as information propagates from layer to layer, as well as overall changes in distributional dynamics at the network level.  This fine-grained analysis revealed significant dynamic range fluctuations and a ``distributional mismatch" between channelwise and layerwise distributions in DWSCNNs that lead to increasing quantized degradation and distributional shift during information propagation.  Furthermore, analysis of the activation quantization errors show that there is greater quantization error accumulation in DWSCNN compared to regular CNNs.  The hope is that such insights can lead to innovative strategies for reducing such distributional dynamics changes and improve post-training quantization for mobile.
\end{abstract}

\vspace{-0.2in}
\section{Introduction}

\begin{figure}[h]
\vspace{-0.2in}
    \centerline{\includegraphics[width=0.4\textwidth]{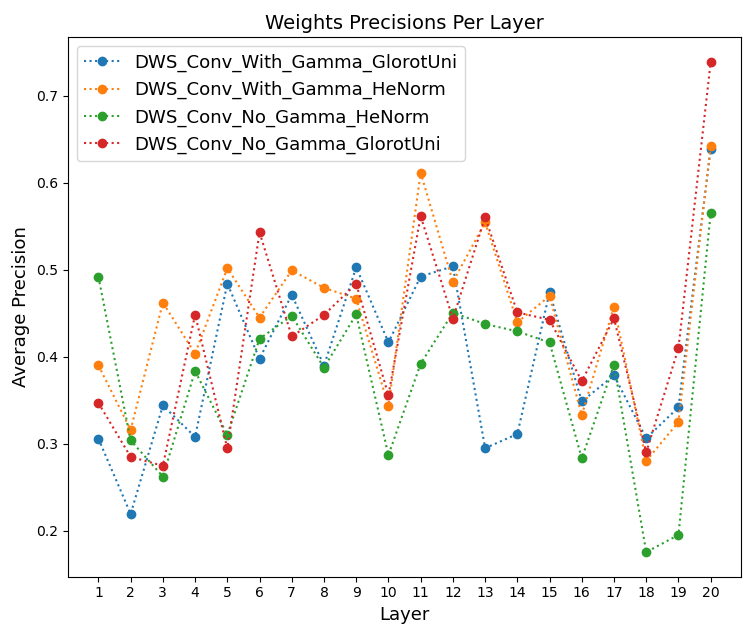}}
    \caption{\footnotesize{}Layerwise average precison (see Eq.~\ref{eq:averageprecision}) of trained weights in the DWS-Convnets. Low average precision at depthwise-conv layers indicate a mismatch between each individual channel's dynamic range and the entire tensor's range. The resulting distributional dynamics lead to significantly higher accumulation of quantization errors. We see a similar pattern in the batchnorm-folded weights and activations of DWSCNNs.}
    \label{fig:dws_wts_precision}
    \vspace{-0.1in}
\end{figure}

\begin{figure*}[t]
\vspace{-0.12in}
    \centering
    \begin{minipage}{0.31\textwidth}
        \centering
        \includegraphics[width=0.95\textwidth]{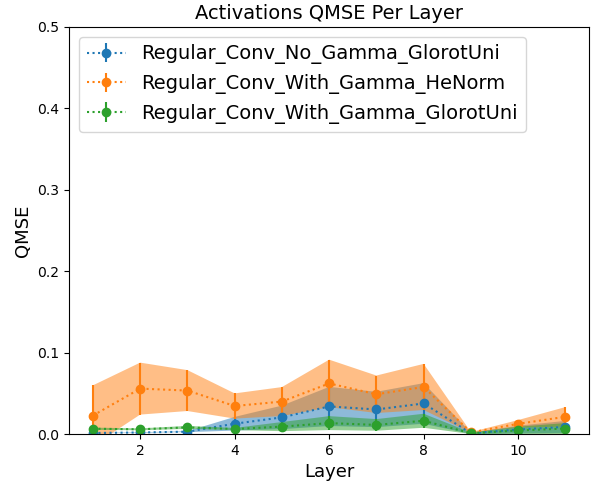} 
    \end{minipage}\hfill
    \begin{minipage}{0.31\textwidth}
        \centering
        \includegraphics[width=0.95\textwidth]{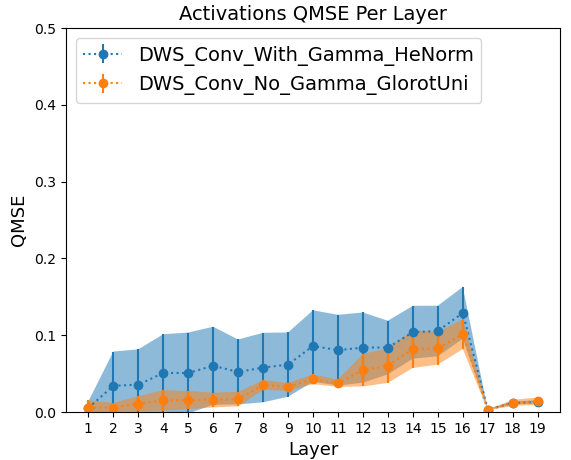}
    \end{minipage}\hfill
    \begin{minipage}{0.31\textwidth}
        \centering
        \includegraphics[width=0.95\textwidth]{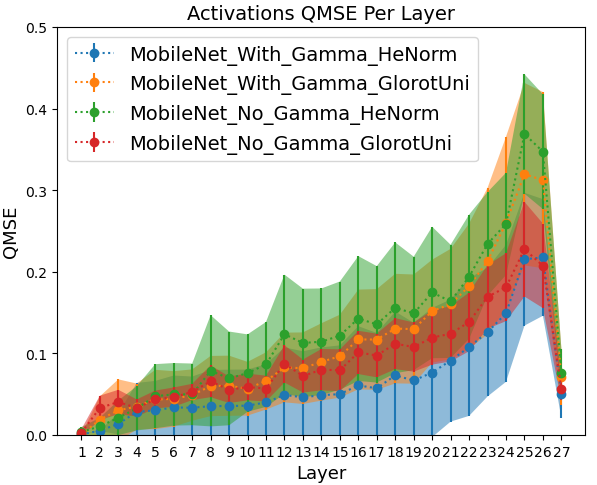} 
    \end{minipage}
    \caption{\footnotesize{} Comparing layerwise QMSE of Regular-ConvNets (left), DWS-ConvNets (center), and MobileNets-V1 (right). We are able to see how depthwise separable convolutional networks accumulate much more QMSE on average when traversing through a quantized DWSCNN. \textbf{Note}: Some outliers were removed from the Regular-ConvNet and DWS-ConvNet plots as they were on a much larger scale and obscure the rest of the results. See Appendix~\ref{appendix} for diagrams of all layerwise QMSE results. The solid line represents the average values across 5 quantization trials and the shaded region is the standard deviation.}
    \vspace{-0.05in}
\label{fig:compare_qmse}
\end{figure*}

The past decade has seen an enormous boom in deep learning research, particularly in the fields of NLP and computer vision. As a result, deep learning algorithms such as convolutional neural network (CNN) models have become more accessible than ever. Mobile devices have become a primary platform on which CNNs have rapidly proliferated. ``AI on-the-edge," has driven an increasing demand for deploying fast, power-efficient CNNs that can maintain highly accurate performance while operating in a resource-constrained environment. Consequently, various avenues of research have looked at making CNNs efficient enough to ``fit" on mobile devices. Methods such as efficient CNN architecture design~\cite{mbnetv1, mbnetv2, mbnetv3, squeezenext, squeezenet, shufflenet, yun2020factorizenet}, weight pruning~\cite{han2015deep_compression, liu_sparsity_2015, yang2020deephoyer, louizos_l0_sparsity, li_filter_prune}, and quantization~\cite{tf_quantize, han2015deep_compression, nagel_adaround, nagel_dfq} are all aimed at reducing the storage, computation, and memory requirements of CNN algorithms. Among these methods, depthwise separable convolutional networks (DWSCNN) such as in MobileNets~\cite{mbnetv1, mbnetv2, mbnetv3} and fixed point integer/quantized inference have become ubiquitous tools for designing efficient ``Mobile AI" algorithms.

As Mobile AI continues to proliferate, so does the need to better understand the behaviour of the models we deploy. Despite the success of MobileNets, there has been a well-documented phenomenon wherein simple post-training static quantization completely destroys their accuracy. While several works have designed solutions to address this, there is still limited data in the literature demonstrating \textit{why} MobileNets quantize so poorly compared to other CNN architectures. Furthermore, we wonder if this problem is inherent to all DWSCNN architectures. 

To investigate this, we do our best to recreate MobileNets-V1 training procedure described in ~\cite{mbnetv1}, with some modifications for the CIFAR-10 experiments. Furthermore, based on findings in~\cite{yun2020begin}, it is possible that choices such as the random weight initialization method and potentially even the use of BatchNorm with/without scaling (ie. the $\gamma$ parameter described in~\cite{batchnorm}) could have significant impact on the trained layerwise distributions to be quantized. Thus, we perform a systematic, iterative series of experiments to isolate the impact of these factors on quantized MobileNets-V1 performance. This experiment is then repeated on a set of smaller DWSCNN architectures (that we will refer to as DWS-ConvNets) to investigate if this problem is inherent to depthwise-separable convolutional networks in general. For a baseline comparison, we also train a corresponding set of Regular-ConvNets. Our systematic ablation study is coupled with fine-grained, layerwise analysis similar to those described in~\cite{yun2020begin, yun2020factorizenet}. 

We find that significantly fluctuating dynamic ranges from layer-to-layer and a ``distributional mismatch" between channelwise and layerwise distributions leads to increasing quantized degradation. Consequently, our fine-grained analysis shows that the quantized mean squared error (QMSE), quantized cross-entropy (QCE), and quantized KL-Divergence (QKL-Div) (defined in Sec.~\ref{sec:results}) of each layer's activations accumulate much more during forward propagation of quantized DWSCNNs and lead to noticeably larger degradation compared to regular CNNs. Thus, indicating that there is greater error propagation and shifting distributional dynamics as information propagates through each layer. Furthermore, DWSCNNs appear to have much larger variation in quantization behaviour depending on the method of random weight initialization used. These observed phenomena would explain why channelwise quantization~\cite{tf_whitepaper} and methods such as~\cite{nagel_dfq, meller_same_same, kurtosis_reg_quant} that decrease distributional mismatch can provide impressive improvements on MobileNets quantization. 

Utilizing fine-grained analysis enables a detailed view of the multi-scale distributional dynamics of our CNN architectures. Thus, facilitating better understanding of how CNN design choices affect the final trained distributions of weights and activations and the complex interactions between each layer's feature mappings and quantization noise. Tracking the layerwise QMSE captures the spatial-channel structure of accumulating quantized activation errors and helps us understand how these errors propagate through the network. Meanwhile, QCE/QKL-Div quantifies the degree of distributional shift and change in distribution dynamics of a network's representations when quantization noise is introduced. Analyzed together, we can gain a deeper understanding of the complex system dynamics involved in CNN quantization. We hope that these insights can lead to further innovative strategies for reducing and compensating for such distributional dynamics changes and improve post-training quantization for mobile deployment.

\section{Motivations and Related Work}
\subsection{Efficient CNN Architectures via Depthwise Separable Convolutions}
Efficient CNN architecture design is now a well-established field with several works~\cite{mbnetv1, mbnetv2, shufflenet, squeezenext, squeezenet} proposing various design patterns for factorizing convolution layers and reducing the computational load of inference. A primary tool for reducing the number of parameters and multiply-accumulate (MAC) operations in a CNN is depthwise-separable (DWS) convolution. Architectures such as MobileNets~\cite{mbnetv1, mbnetv2, mbnetv3}, ShuffleNets~\cite{shufflenet} and FBNets~\cite{fbnets} make heavy use of DWS convolution to achieve state-of-the-art accuracy for low-power/efficient CNNs. Depthwise separable convolution factorizes regular or ``dense" convolution into a $K\times K$ depthwise convolution (that is, each convolutional kernel of size $K$ is only applied to a single input channel) for low-dimensional feature extraction and a ``pointwise" convolution (a dense convolution with $1\times1$ kernels) for mixing channel information. As mentioned in~\cite{mbnetv1}, this leads to between 8 to 9 times reduction in MACs for a $3\times3$ DWS convolution compared to its regular/dense counterpart. Furthermore, with publicly available ImageNet-trained checkpoints of the MobileNets ``model family," their efficient, generalized features can be leveraged for various application-specific tasks via transfer learning. In this way they can be used ``off-the-shelf" for finetuning with much less training overhead compared to training accurate models from scratch. However, even though MobileNets can already run incredibly fast with floating point (fp32) operations on a mobile CPU, further storage and power reductions can be gained if they are converted for 8-bit fixed point integer (quint8) operation and run on low-power, parallelized hardware such as a digital signal processor (DSP). Consequently, creating robust models that maintain accuracy during quantized inference has been a growing area of research with particular interest in improving the robustness of compact models like MobileNets.

\begin{figure*}[t]
\centerline{\includegraphics[width=13cm]{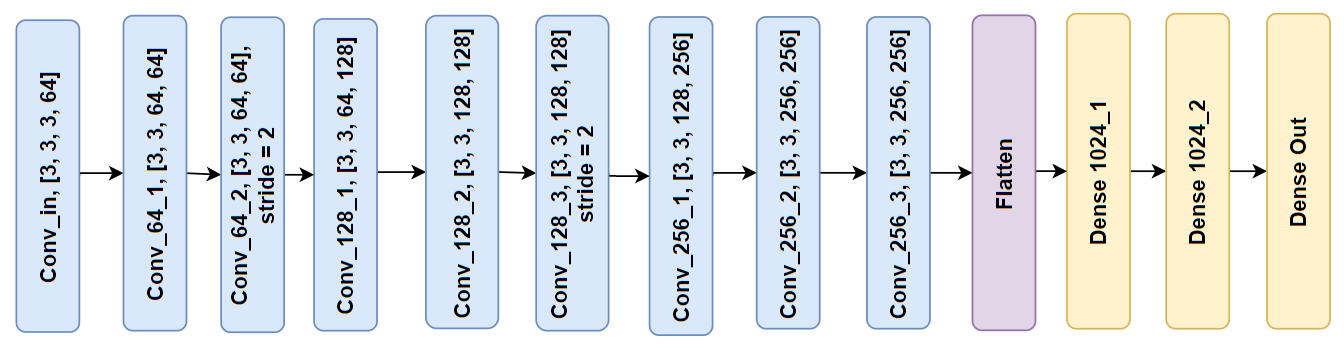}}
\caption{\footnotesize{}\textbf{Simple ToyNet Macroarchitecture}. For our ablation study we define a simple macroarchitecture (ie. input/output channels of each layer, stride, kernel size). We then ablate through a few hyperparameter settings that might affect the final layerwise distributions. Shape of weight tensors for regular convolution is in square brackets. For DWS-ConvNets, we replace regular convolution with DWS convolution while preserving input/output dimensions.}
\label{fig:toynet}
\end{figure*}

\subsection{Fixed-point Quantization For Efficient Mobile Inference}

In conjunction with efficient CNN architecture design, low bit-width (16 bits and below, though most commonly 8 bits), fixed point quantization has enabled highly parallelized processors such as DSPs to run fast, low-power inference entirely with integer arithmetic. These methods \cite{tf_quantize, log_quant, vector_quant} project the neural network weights and activations of each layer onto a low-dimensional, discretized space while minimizing loss of information. However, the noise induced by quantization error has complex interactions with the weights and activations of each layer and its impact on CNN output behaviour can be difficult to quantify. Thus, it can often be hard to predict which CNN architectures will quantize well (ie.  are ``quantization friendly") and are suitable for deployment.

Given the problem of quantization robustness, various research works \cite{tf_quantize, tqt, nagel_dfq} explore methods to increase the robustness of models to quantization noise. Methods such as quantization-aware training (QAT)~\cite{tf_quantize} and trained quantization thresholds (TQT)~\cite{tqt} make use of simulated quantization and the straight-through estimator (STE) to train the network to adapt to  quantization noise. Note that since these methods simulate quantization, the training is bit-width specific and a network must be retrained if one were to adapt a model for a different bit-width. Furthermore, quantization training can require hyperparameter tuning of its own, thus extending the design cycle. Oftentimes, the choice of how to optimize a model for quantization is based on available tools and trial-and-error. While many of the devised ``quantization fixes" have shown remarkable results \cite{nagel_dfq, nagel_adaround, tqt}, they aren't necessarily guaranteed to transfer across applications. For example, image reconstruction and other continuous value prediction/regression tasks may have a much lower error tolerance than classification. Thus, we seek to use a systematic approach which can help delineate the various factors affecting quantization robustness. In this way, we can make more informed choices on how to improve the quantized behaviour of our CNN models.

\cite{quant_friendly_dws, meller_same_same} perform a layerwise analysis of the signal-to-quantization-noise-ratio (SQNR) in a CNN. They use SQNR to estimate the amount of useful information passing from layer to layer in a quantized CNN and seek to maximize it. \cite{quant_friendly_dws} analyzes layerwise SQNR to identify architectural choices that were hurting the quantized performance of MobileNets-V1. Similarly, we would like to use layerwise analysis to better understand the poor quantization of MobileNets and other DWSCNN models.

In a newer area of ``robust quantization" works, the authors of \cite{l1_grad_reg_quant, kurtosis_reg_quant} seek to train models that are generally robust to quantization noise (using $L_1$ gradient regularization and a kurtosis-based weight regularizer (KURE) respectively) such that they can be easily quantized at varying bit-widths using simple post-training quantization methods. These kinds of models are trained to be robust to a broad range of perturbations that may be induced by various uniform quantization settings such that they can be smoothly deployed to quantized inference environments without further, potentially costly, quantization-based optimizations. Drawing on the ideas of \cite{quant_friendly_dws, l1_grad_reg_quant, kurtosis_reg_quant} we hope to gain better insight on how different parametrizations of CNN can be constructed that tend towards learning quantization-robust solutions. In our work, we will focus on 8-bit uniform quantization described in \cite{tf_quantize} as it has become the most common method adopted in industry for mobile devices.

\section{Experiments}
\label{sec:experiment}

\begin{table*}
	\setlength{\tabcolsep}{6pt}
	\centering
	\resizebox{1.01\textwidth}{!}{%
		\csvreader[tabular=|l|c|c|c|c|c|c|,
        table head=\hline \textbf{Network Architecture} & \textbf{FP32 Acc (\%)} & \textbf{QUINT8 Acc (\%)} & \textbf{QMSE} & \textbf{QCE} & \textbf{QKL-Div} & \textbf{Percent Acc Decrease}\\\hline,
        late after line=\\\hline]%
        {updated_regular_and_mbnet_mean_quant_results.csv}{Network Architecture=\Network, FP32 Acc=\FP, QUINT8 Acc=\QINT, QMSE=\QMSE, QCE=\QCE, QKL-Div=\QKLD, Percent Acc Decrease=\Percent}%
        {\Network & \FP & \QINT & \QMSE & \QCE & \QKLD & \Percent}%
		}
		\caption{Detailed quantization results for CIFAR-10 networks. Quantization results reported as mean and standard deviation across five quantization trials. The output QMSE, QCE and QKL-Div of DWS convolution based networks are noticeably higher than regular CNNs.}
	\label{cifar_quant_results}
\end{table*}

\begin{table*}
	\setlength{\tabcolsep}{6pt}
	\centering
	\resizebox{1.01\textwidth}{!}{%
		\csvreader[tabular=|l|c|c|c|c|c|c|,
        table head=\hline \textbf{Network Architecture} & \textbf{FP32 Acc (\%)} & \textbf{QUINT8 Acc (\%)} & \textbf{QMSE} & \textbf{QCE} & \textbf{QKL-Div} & \textbf{Percent Acc Decrease}\\\hline,
        late after line=\\\hline]%
        {ImageNet_quant_results.csv}{Network Architecture=\Network, FP32 Acc=\FP, QUINT8 Acc=\QINT, QMSE=\QMSE, QCE=\QCE, QKL-Div=\QKLD, Percent Acc Decrease=\Percent}%
        {\Network & \FP & \QINT & \QMSE & \QCE & \QKLD & \Percent}%
		}
		\caption{Detailed quantization results for VGG-19 and MobileNet-V1-1.0-224 trained on ImageNet. Quantization results reported as mean and standard deviation across three quantization trials. While the scale of QMSE is different from CIFAR-10 due to the softmax distribution being computed over 1000 classes, we observe similar trends of greater accumulated error and distributional shift in MobileNet.}
	\label{imgnet_quant_results}
\end{table*}

Since we want to explore quantization results of both ``official" MobileNets architecture and DWSCNNs in general, we run multiple trainings with the CIFAR-10 dataset. In this study, the detailed multi-scale distribution dynamics analysis on the variety of quantized network architectures was conducted on CIFAR-10 as its smaller size allows for a broader ablation study given resource constraints.  Thus, we can quickly do a systematic comparison across multiple DWS convolution based architectures. Due to the much smaller $32\times32$ images of CIFAR-10, we could not use the exact same MobileNets-V1 architecture as reported in~\cite{mbnetv1}. The main architectural differences are as follows:
\begin{itemize}
    \item We do not downsample with stride = 2 at the first DWS-Conv block with output channels 128\vspace{-0.06in}
    \item We do not downsample with stride = 2 at the first DWS-Conv block with output channels 1024\vspace{-0.06in}
    \item Thus, the input tensor shape to Global Average Pooling layer is $4\times4\times1024$ rather than $1\times1\times1024$
\end{itemize}

In addition to directly comparing regular convolution vs. depthwise separable convolution, we also performed additional experiments with ResNet-34 using the same set of experiment conditions (ie. training hyperparameters, weight initializer choices, use of BatchNorm $\gamma$-scaling). For ResNet-34, the CIFAR-10 specific architectural modifications, quantization results, and layerwise plots can be found in Appendix~\ref{appendix}. We refer readers to~\cite{mbnetv1, resnet} for original architecture details. We tried to preserve some of the overall topologies of MobileNets-V1 (eg. downsampling after the first convolution layer, downsampling when output channels increase etc.) while creating an architecture that could get reasonable results. We tried our best to recreate the training process of MobileNets. All experiments were conducted using Tensorflow 1.15. The optimizer, hyperparameters, data augmentations etc. are as follows:
\begin{itemize}
    \item RMSProp. Momentum and decay = 0.9, epsilon = 1.0\vspace{-0.06in}
    \item Training batch size = 128\vspace{-0.06in}
    \item Logging and quantization batch size = 800\vspace{-0.06in}
    \item Number of epochs = 200\vspace{-0.06in}
    \item Initial learning rate 0.045. Scaled by 0.97 each epoch.\vspace{-0.06in}
    \item Data augmentation: random rotation, width and height shift, zoom, random horizontal and vertical flip
\end{itemize}

While the macroarchitecture and the above listed hyperparameters were fixed, we also tried comparing the effect of weight initializaton method (Glorot Uniform~\cite{glorot} vs He Normal~\cite{he_init}) and use of BatchNorm scaling (ie. we compared applying BatchNorm with and without $\gamma$ parameter). In~\cite{yun2020begin}, the authors show that BatchNorm and weight initialization methods can have a significant effect on quantization results and so we wanted to observe the effects they may have had on MobileNets-V1. Some hyperparameter choices were slightly adapted for CIFAR-10 such as number of training epochs, learning rate decay schedule etc. We decided to train for longer and decay the learning rate a little bit slower based on similar training setups for CIFAR-10 experiments such as in \cite{wide_resnet, resnet}. For the DWS-ConvNets and Regular-ConvNets, we use the architecture described in Figure~\ref{fig:toynet}. Dropout is used in between all dense layers with a keep-probability of 0.5. Once trained, we follow the model analysis described in~\cite{yun2020begin, yun2020factorizenet} and observe the statistics of the trained weights, activations, and BatchNorm folded (BN-Fold) weights\footnote{Where $\gamma$ is batchnorm scaling parameter, $w$ and $w_{fold}$ are weights and BN-Fold weights respectively, $EMA(\sigma^2_B)$ is variance statistics of the given layer collected during training, $\epsilon$ is a small constant for numerical stability} of each layer (see Eq.~\ref{eq:bn_fold}). Thus, the logging and quantization batch size refers to the batch size used for computing the activation stats. 

\begin{equation}
\label{eq:bn_fold}
w_{fold} = \frac{\gamma w}{\sqrt{EMA(\sigma^2_B) + \epsilon}}
\end{equation}

Additionally, we also define QMSE, QCE, and QKL-Div as the mean squared error, cross-entropy, and KL-Divergence between the fp32 model/hidden-layer outputs and the dequantized quint8 model/hidden-layer outputs respectively. QMSE quantifies the average distance/error whereas QCE and QKL-Div are able to capture differences in the output distribution shapes. Thus, by analyzing these hidden layer output and model output statistics, we can observe the accumulated quantization errors as well as the shifting distribution dynamics at different scales. Note, the QCE/QKL-Div of hidden layers is computed based on the distribution of activations aggregated over an entire batch whereas the QCE/QKL-Div of the model output is computed as the mean cross-entropy/KL-divergence between the softmax output of the fp32 model and quint8 model (ie. the same mean cross-entropy loss computation that would be performed for classification training). We use natural log for all entropy calculations so as to match and compare with the cross-entropy loss observed during training. Thus, for the layerwise quantization error analysis, QCE/QKL-Div illustrate the distributional dynamics of each layer's representations under quantization and how they might change during quantized information propagation through each layer. To compute QCE/QKL-Div, we collect histograms of each layer's activations for the quint8 and fp32 model. The approximate discrete distributions are then used for computing QCE and QKL-Div.

\begin{equation}
\label{eq:averageprecision}
average\_precision = \frac{1}{K}\sum_{i=1}^K \frac{range_i}{range_{tensor}}\
\end{equation}

For layerwise statistics, we are primarily concerned with the range and average precision\footnote{Where $range_i$ is range of $channel_i$ of convolution weights, $range_{tensor}$ is range of convolution weight tensor, $K$ is number of channels/filters} of each layer (see Eq.~\ref{eq:averageprecision}, also defined in ~\cite{nagel_dfq}). For activations, we perform percentile clipping to get the range. However, this percentile clipping slightly differs from true percentile clipping. Instead, we follow the method in Tensorflow Graph Transform tool~\cite{tf_graph_transform} for min/max percentiles since we use it for creating quantized inference graphs. We use min/max percentile of 5\% for both stats logging and activation quantization during inference.

For quantization, we perform 5 trials with randomly sampled training batches for calibrating activation ranges. This is to measure the variation in quantized performance caused by different calibration sets. For each trial, we record the activation statistics of each layer\footnote{For activations, we can directly access range from the quantized graph} to see how these distributions change and how they might correlate with quantization accuracy. 

\vspace{-0.1in}
\section{Results}
\label{sec:results}
\vspace{-0.1in}
\begin{figure*}

    \centering
    \begin{minipage}{0.31\textwidth}
        \centering
        \includegraphics[width=0.98\textwidth]{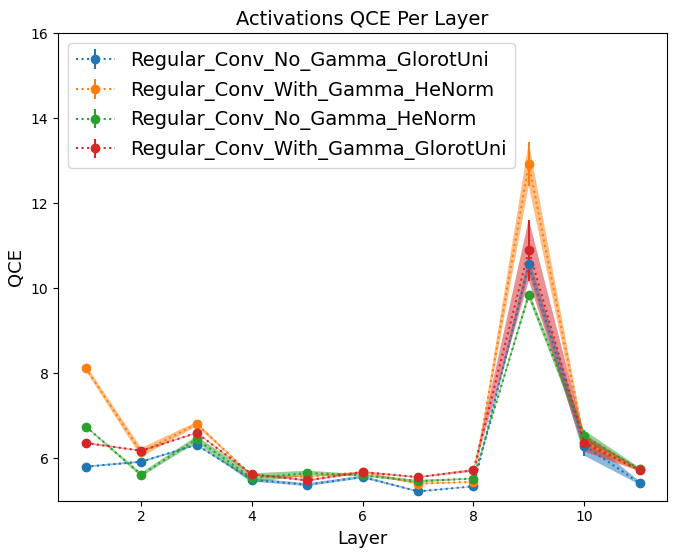}
    \end{minipage}\hfill
    \begin{minipage}{0.31\textwidth}
        \centering
        \includegraphics[width=0.95\textwidth]{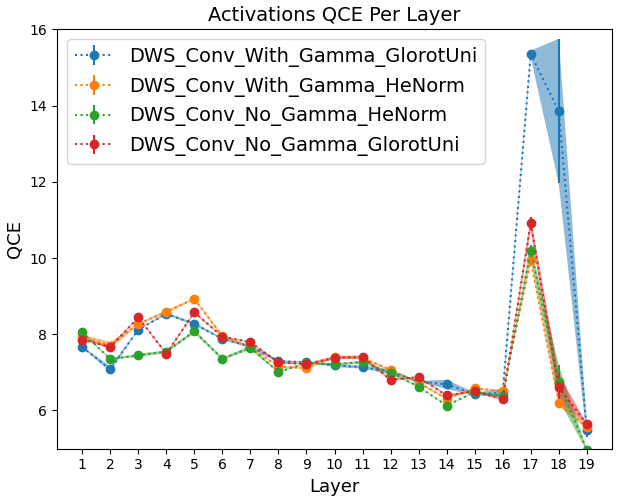}
    \end{minipage}\hfill
    \begin{minipage}{0.31\textwidth}
        \centering
        \includegraphics[width=0.97\textwidth]{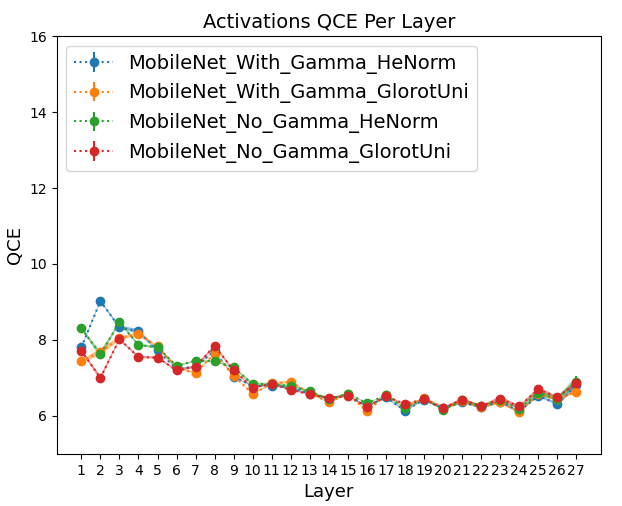} 
    \end{minipage}
    \begin{minipage}{0.31\textwidth}
        \centering
        \includegraphics[width=0.93\textwidth]{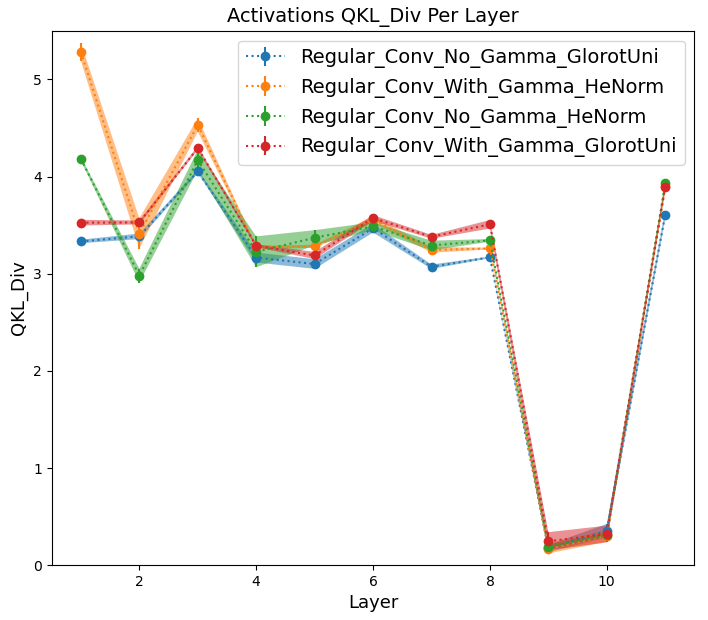} 
    \end{minipage}\hfill
    \begin{minipage}{0.31\textwidth}
        \centering
        \includegraphics[width=0.98\textwidth]{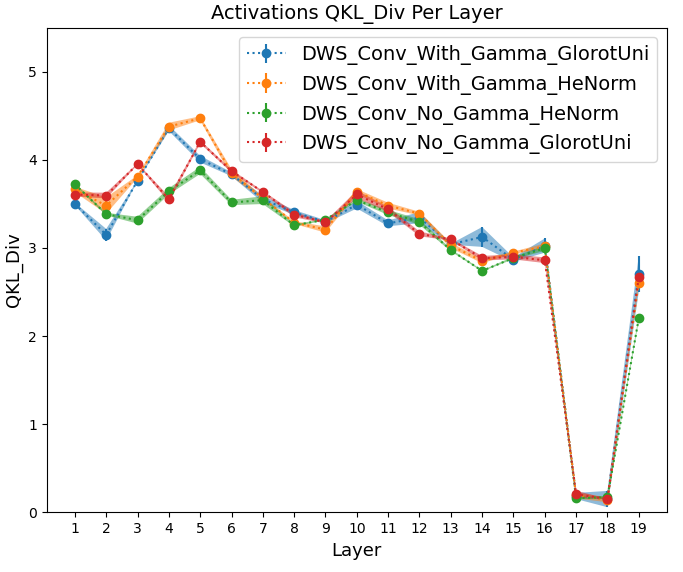}
    \end{minipage}\hfill
    \begin{minipage}{0.33\textwidth}
        \centering
        \includegraphics[width=0.98\textwidth]{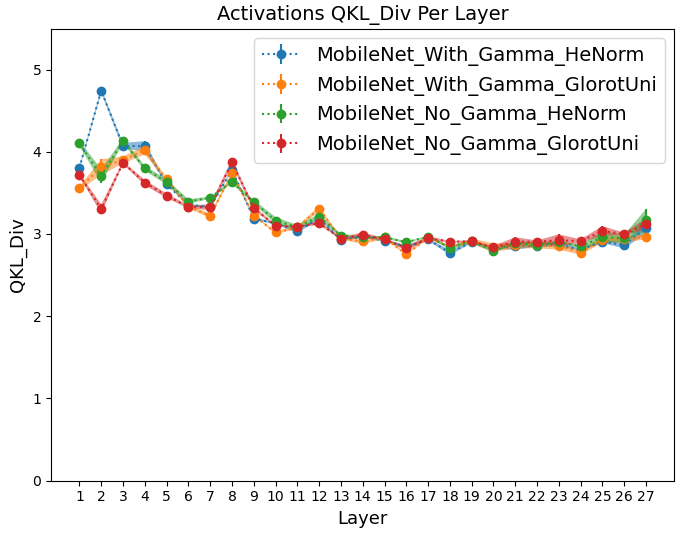} 
    \end{minipage}
    \caption{\footnotesize{} Layerwise QCE (top row) and QKL-Div (bottom row) for all trained networks. Regular CNN (left), DWS-ConvNet (center), and MobileNets-V1 (right). The drop in QKL-Div might explain why, despite significantly accumulatd QMSE, some networks are able to recover a relatively lower QMSE. The solid line represents the average values across 5 quantization trials and the shaded region is the standard devation.}
    
\label{fig:all_quant_dist_dynamics}
\end{figure*}

In Table~\ref{cifar_quant_results} we compare a few different testing metrics for the trained models. In terms of accuracy, we look at floating point accuracy, quantized accuracy, and relative/percent accuracy decrease (ie. change in accuracy over fp32 accuracy). As described in Sec.~\ref{sec:experiment}, we also observe the QMSE, QCE, and QKL-Div between the fp32 model output and quint8 model output. We see that on average, MobileNet-V1 and DWS-ConvNets experience much larger degradation in performance under quantization compared to the Regular CNNs. This is demonstrated by the much larger average percent accuracy decrease ($13.0-32.09$ vs. $6.12-9.22$), mean QMSE ($0.0186-0.0592$ vs. $0.00764-0.0170$), mean QCE ($0.77-2.51$ vs. $0.37-0.66$), and mean QKL-Div ($0.45-2.04$ vs. $0.18-0.46$). Furthermore, the depthwise-separable based CNNs seem to have much larger variation in quantization behaviour depending on random weight initialization method. In our ResNet-34 experiments, we observed results similar to the Regular-Conv networks (see Appendix~\ref{appendix}). This is rather interesting since we had initially hypothesized that the residual learning block might enable learning of more compact distributions. Thus, reducing quantization noise/errors.

We now move onto a fine-grained analysis of the distributional dynamics of each network type. Coupled with the results observed in Table~\ref{cifar_quant_results}, we can gain a deeper understanding of how each layer's distributions interact with quantization and the resulting accumulation of quantization errors (QMSE) and distributional shifts  (QCE/QKL-Div). 

\begin{figure*}

    \centering
    \begin{minipage}{0.31\textwidth}
        \centering
        \includegraphics[width=0.98\textwidth]{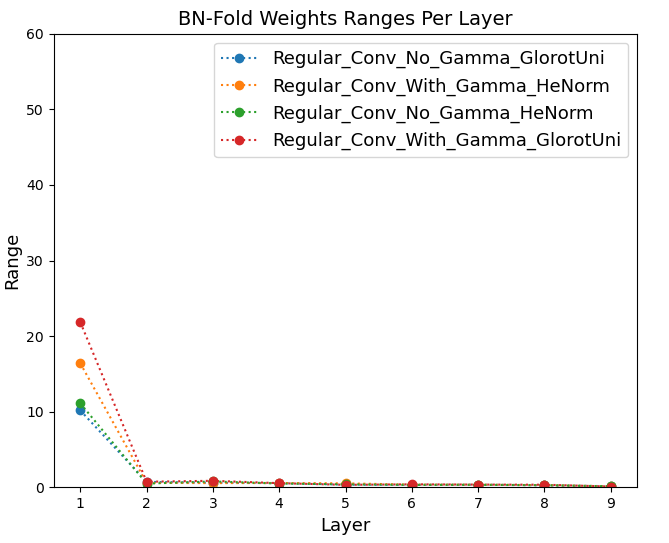} 
    \end{minipage}\hfill
    \begin{minipage}{0.31\textwidth}
        \centering
        \includegraphics[width=0.95\textwidth]{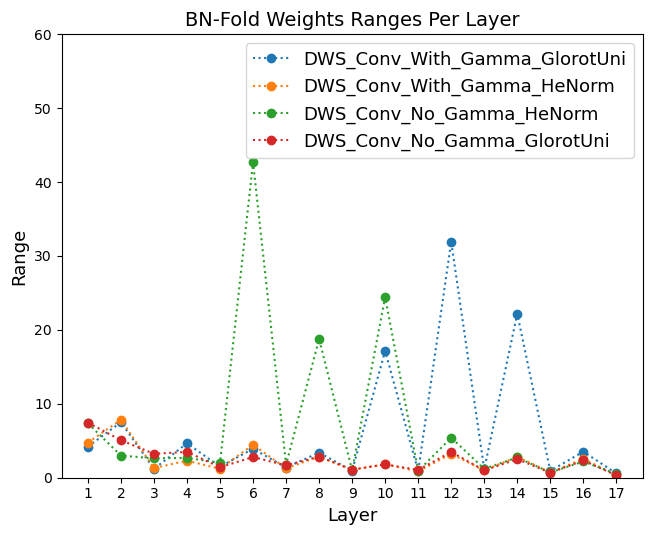}
    \end{minipage}\hfill
    \begin{minipage}{0.31\textwidth}
        \centering
        \includegraphics[width=0.95\textwidth]{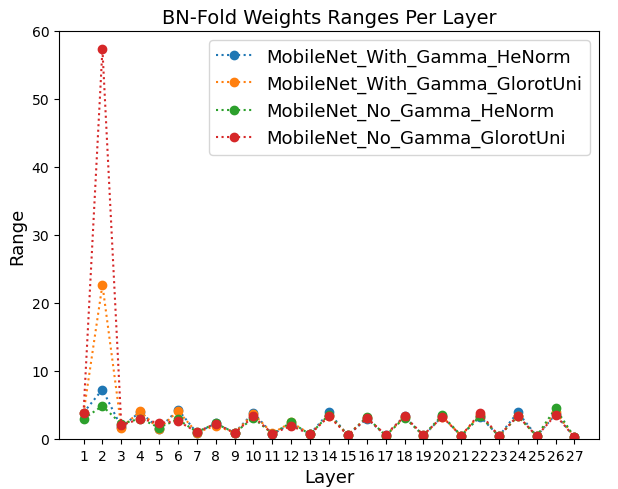} 
    \end{minipage}
    \begin{minipage}{0.31\textwidth}
        \centering
        \includegraphics[width=0.98\textwidth]{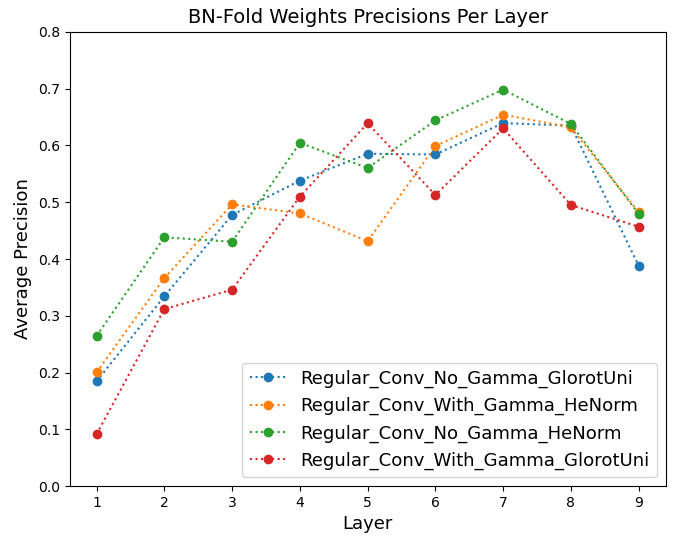} 
    \end{minipage}\hfill
    \begin{minipage}{0.31\textwidth}
        \centering
        \includegraphics[width=0.95\textwidth]{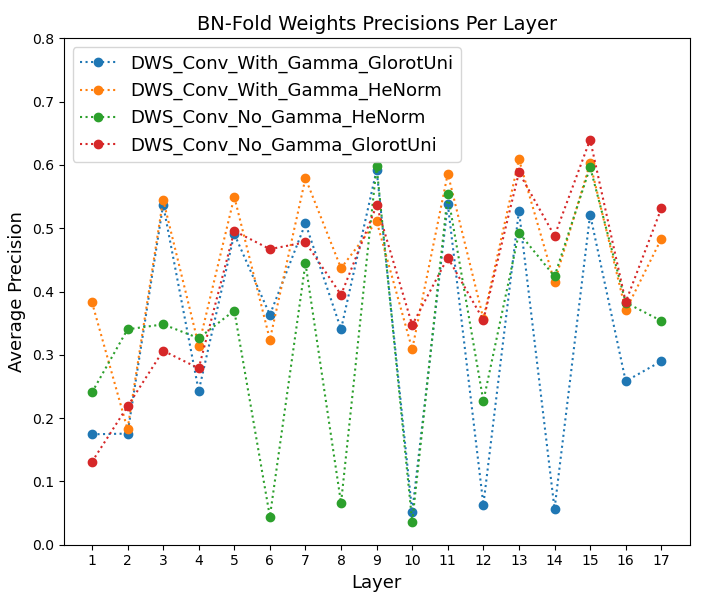}
    \end{minipage}\hfill
    \begin{minipage}{0.31\textwidth}
        \centering
        \includegraphics[width=0.98\textwidth]{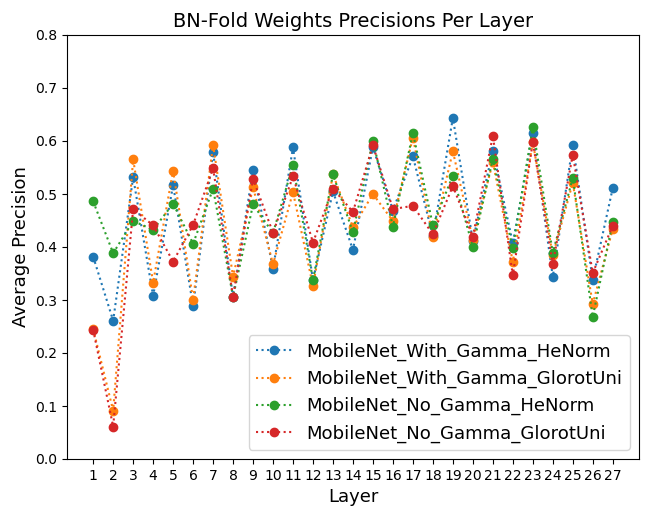} 
    \end{minipage}
    \caption{\footnotesize{} Layerwise BN-folded weights dynamic range (top row) and average precision (bottom row) for all trained networks. Regular CNN (left), DWSCNN (center), and MobileNets-V1 (right). The fluctuating range and average precision help explain the quantization degradation of DWSCNNs.}
    
\label{fig:compare_bn_fold_ranges}
\end{figure*}

\begin{figure*}

    \centering
    \begin{minipage}{0.45\textwidth}
        \centering
        \includegraphics[width=0.93\textwidth]{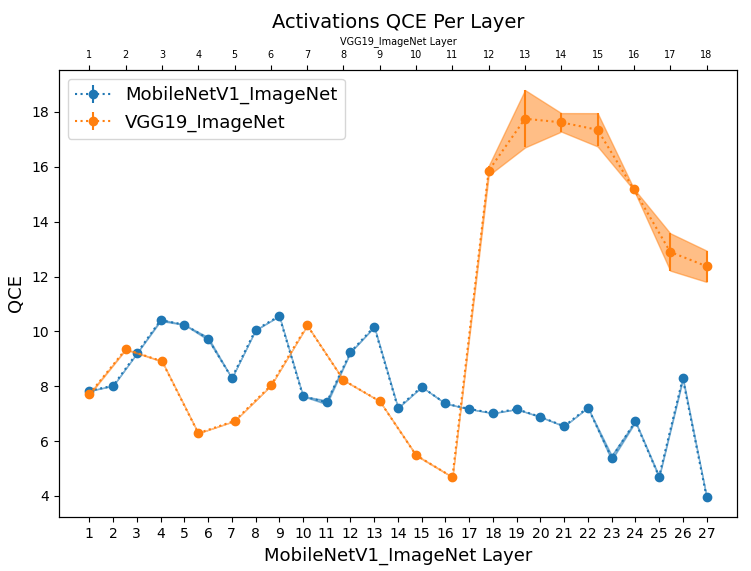}
    \end{minipage}\hfill
    \begin{minipage}{0.45\textwidth}
        \centering
        \includegraphics[width=0.94\textwidth]{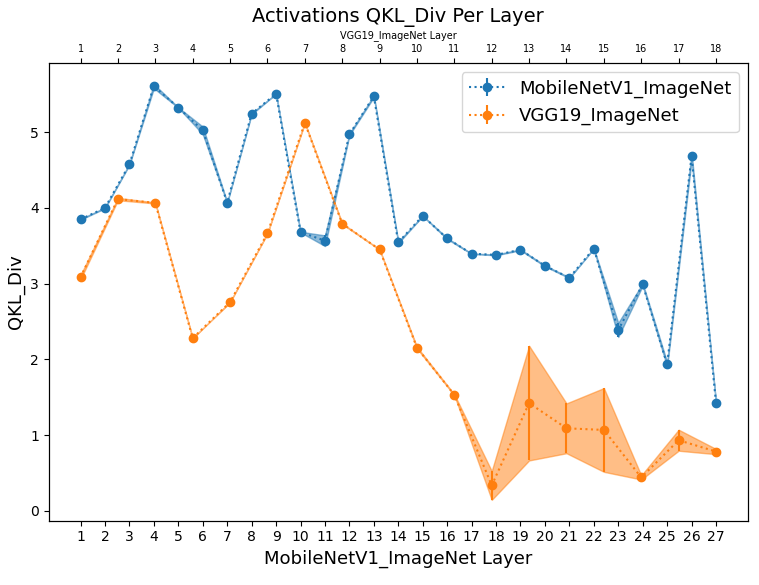}
    \end{minipage}\hfill

    \caption{\footnotesize{} Layerwise quantization of VGG-19 and MobileNet-V1 on ImageNet. Layerwise QCE (left), Layerwise QKL-Div (right).}
    \label{fig:imgnet_quant_stats}
\end{figure*}

\begin{figure*}

    \centering
    \begin{minipage}{0.31\textwidth}
        \centering
        \includegraphics[width=0.95\textwidth]{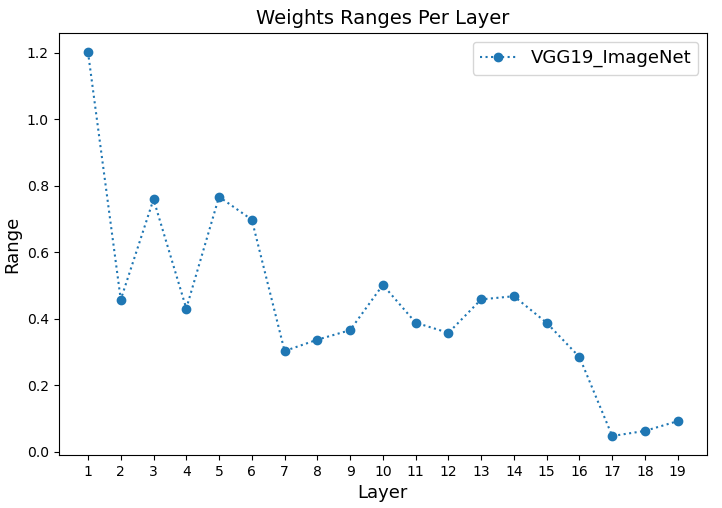} 
    \end{minipage}\hfill
    \begin{minipage}{0.31\textwidth}
        \centering
        \includegraphics[width=0.95\textwidth]{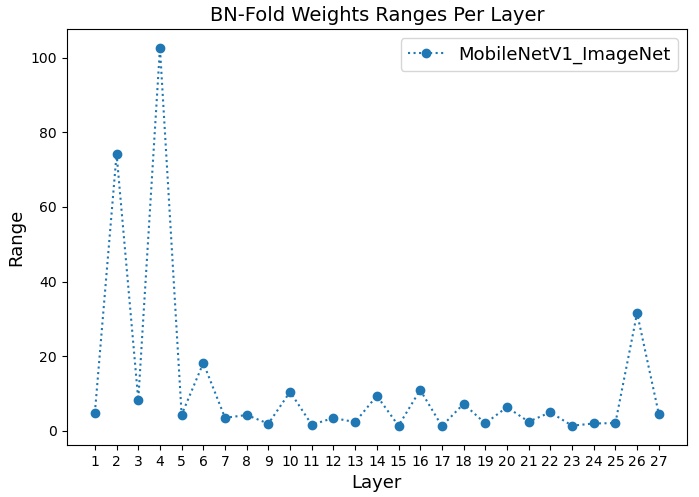}
    \end{minipage}\hfill
    \begin{minipage}{0.31\textwidth}
        \centering
        \includegraphics[width=0.98\textwidth]{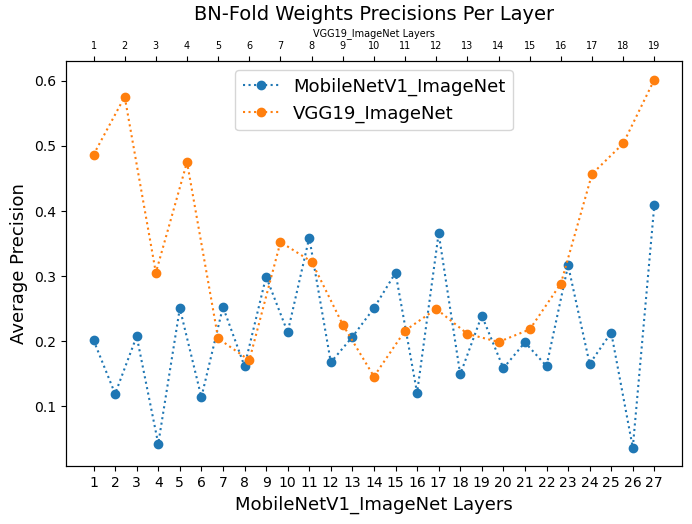} 
    \end{minipage}
    
    \caption{\footnotesize{} Layerwise weights/BN-Folded weights range of VGG-19 (left) and MobileNet-V1 (center) on ImageNet. Layerwise average precision of weights/BN-Folded weights of VGG-19 and  MobileNet-V1 on ImageNet (right). Note the vastly different scales for weights ranges.}
    \label{fig:imgnet_wts}
\end{figure*}

\section{Discussion}
To analyze the layerwise distributions of the different networks, we collected stats on the range and average precisions for the weights, BN-Folded weights, and activations (see Figures~\ref{fig:dws_wts_precision},~\ref{fig:compare_bn_fold_ranges}). For the sake of space, we have omitted most of the plots related to the weights distributions since it is the BN-Folded weights that will get quantized. However, comparing the distributions of weights before and after BN-Folding can reveal interesting insights on how each layer is being scaled and the resulting distributional shift. We also left out the layerwise activations plots as they follow similar trends to the BN-Folded weights. Detailed results and figures are included in Appendix~\ref{appendix}. As mentioned in \cite{yun2020factorizenet}, the average precision gives a measure of how well the layerwise quantization encodings represent the information in an individual channel. If the precision is low, it is possible that the quantization noise may ``wash out" the information of the given individual channel. Thus, range and average precision capture information about the distributions at both a layerwise and channelwise scale. Together, they give a picture of the magnitude of quantization noise, and the potential interactions of information in individual channels with the quantization noise. 

We can see in the plots that the DWS convolution based architectures have significantly fluctuating ranges and average precisions. The sharply fluctuating plots illustrate the poorly behaved distributions of weights and activations that DWSCNNs tend to learn. Most notably, at depthwise convolution layers (even-numbered layers in the plots), the dynamic range peaks while average precision simultaneously drops. Suggesting that features extracted by the depthwise convolution will have significantly more quantization noise compared to their regular convolution counterparts. Low average precision suggests that the channelwise connection sparsity of depthwise-separable convolutions could be a detriment to quantized behaviour, possibly leading to learned distributions with low inter-channel correlation. Thus, causing tensorwise quantization to be a non-representative mapping of the weights and activations into discretized space.

Besides examining the layerwise distributions, we can directly look at our layerwise quantization noise statistics to better understand how the observed distributional dynamics manifest in the quantized behaviour. In Figures~\ref{fig:compare_qmse},~\ref{fig:all_quant_dist_dynamics} we can observe the interactions between each hidden layer's output activations and the noise induced by uniform 8-bit quantization. It is immediately apparent that QMSE accumulates significantly more in the DWS-ConvNets and MobileNets-v1\footnote{As noted in Figure~\ref{fig:compare_qmse}, some outliers were removed from the QMSE plots. Full layerwise QMSE plots in supplementary material}. While QMSE captures the 2D-structure of the accumulated quantization error, QCE and QKL-Div describe the distributional shift induced by quantization and that accumulation during information propagation. From an information theoretic perspective, we could interpret them together as the amount of information in the hidden representations and how well the 8-bit encoding is representing that information. Considering the DWS-ConvNet and Regular-ConvNet plots, the peaking QCE that coincides with a drop in QKL-Div could be interpreted as maximal information being transmitted by later layers (entropy of the distribution) and a fairly representative mapping of the hidden distribution from continuous fp32 space to discrete quint8 space. This drop in QKL-Div might explain why despite relatively high peaks in QMSE, most of the DWS networks can ``recover" and return to a lower QMSE at the output. The drop in QKL-Div would imply that the overall distribution of activations is preserved and consequently the relevant information is still passed onto the following layers. By contrast, the quantization noise statistics in MobileNet-v1 demonstrate a steadier level of distributional shift likely due to the deeper architecture. However, MobileNets still suffers from fluctuating layer dynamics and distributional mismatch leading to an overall larger accumulation of QMSE. Thus, it is not able to quantize as well as the Regular-ConvNets.

Based on these insights, we believe there are a few approaches that could be explored to improve quantization by either reducing accumulation of errors or improving the alignment of layerwise and channelwise distributions. Minimizing layerwise QMSE in addition to quantized output errors could be beneficial at lower bit-widths (eg. 4-bit, 2-bit quantization) where the backpropagation of STE's biased gradient estimate from the output layer can lead to increasingly inaccurate gradient estimates. The QMSE loss can act as a closer, more accurate gradient feedback, especially for earlier layers. With respect to minimizing misalignment of distributions, finetuning with weight normalization or a regularizer that better aligns the weight distributions could help reduce quantization distributional shifts and make the layer/tensorwise quantization mapping more representative of the information in each individual channel.

\section{ImageNet Analysis}
To study the distributional dynamics on more complex deep neural network architectures and on a much larger dataset, we conduct a similar analysis on VGG-19 and MobileNet-V1 trained on ImageNet\footnote{We leveraged ImageNet-trained fp32 checkpoints of VGG-19 and MobileNetV1-1.0-224 from \cite{google_models}.}.  We follow the same quantization procedures as described in Section~\ref{sec:experiment}. However, due to time and compute constraints, we limited the number of quantization trials conducted to 3. We also increase the size of the quantization/activation logging set to 2000. For layerwise quantization and activation stats, we iterate through 2000 images with batches of 50 images and compute the mean due to RAM constraints.

Table~\ref{imgnet_quant_results} shows the quantized output results. As expected, the quantized accuracy drop for MobileNet is catastrophic. However, the relative degradation of VGG-19 ($8.58\%$ relative accuracy decrease) stays fairly consistent with CIFAR-10 results. Upon comparing the range and average precision of the BN-Folded weights\footnote{The VGG-19 checkpoint does not use BatchNorm. Thus we compared VGG-19 weights to MobileNet BN-Folded weights.} of the two networks (see Figure~\ref{fig:imgnet_wts}) we see that the issue of mismatched distributions combined with large, spiking dynamic ranges is even more pronounced in MobileNet trained on ImageNet. As DWS convolution decouples the convolutional channels from each other, we believe the diversity of ImageNet data significantly aggravates the misalignment of channelwise distributions vs. layerwise distributions and as a result leads to a significant loss of information. While VGG-19 also has a decrease in average weight precision near the ``middle" of the network, its dynamic ranges are about an order of magnitude smaller and the average precision is still generally better.

We see in Figure~\ref{fig:imgnet_quant_stats} that the higher layerwise QKL-Div for MobileNets indicate much greater quantized distributional shift. Due to differences in preprocessing\footnote{MobileNet scales images to the range $[-1, 1]$ whereas VGG only zero-centers each RGB channel without scaling}, the layerwise QMSE and layerwise activations data are on completely different scales (VGG-19 has much larger ranges). Thus, making it hard to compare these values\footnote{The activations and QMSE plots are included in Appendix~\ref{appendix}}. However, similar to CIFAR-10 results, MobileNet trained on ImageNet had much larger output QMSE, QCE, and QKL-Div. Interestingly, this shows how large dynamic activations range do not automatically translate to poor quantization performance. Instead, accumulated errors, distributional shifts and the multi-scale distributional dynamics of each layer should be observed together as a whole to better understand the quantization dynamics at play.

\section{Conclusion}
We perform a systematic ablation study with fine-grained analysis to understand why DWSCNNs seem to quantize more poorly than regular CNNs on average. We find that this appears to be inherent to depthwise separable convolutions owing to the fact that depthwise convolutions tend to learn representations with greatly fluctuating dynamic ranges and significant intra-layer distributional mismatch. Analyzing CNN quantization through the lens of multi-scale distributional dynamics, we observe that depthwise convolution based CNNs suffer from larger accumulation of quantization errors and distributional shift. We believe these insights point to potential new ways to mitigate such distributional dynamics changes and improve robustness of post-training quantization for efficient depthwise-convolution based CNNs.  As MobileNet-V2 also quantizes poorly, future work involves extending our analysis to the inverted bottleneck residual blocks described in \cite{mbnetv2}. Furthermore, we would also like to try setting certain layers of a quantized network to floating point operation and observing the correlation of layerwise QMSE/QCE/QKL-Div with quantized output behaviour.

{\small
\bibliographystyle{ieee_fullname}
\bibliography{references}
}

\clearpage

\appendix
\section{Appendix}
\label{appendix}
\subsection{Plots From Main CIFAR-10 Ablation Study}
\begin{figure}
    \centering
    \begin{minipage}{0.5\textwidth}
        \centering
        \includegraphics[width=0.98\textwidth]{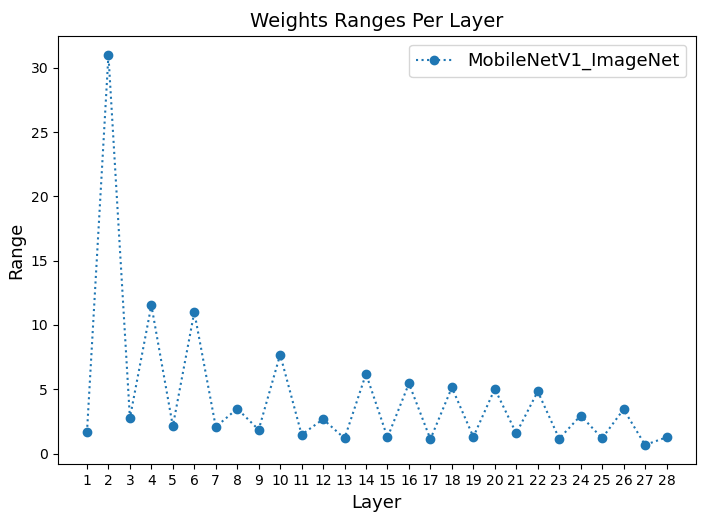} 
    \end{minipage}\hfill
    \begin{minipage}{0.5\textwidth}
        \centering
        \includegraphics[width=0.98\textwidth]{MobileNetV1_ImgNet_BN_Fold_Wts_Ranges} 
    \end{minipage}\hfill

    \caption{\footnotesize{} Layerwise weights range (top) and BN-Folded weights range (bottom). Observing the BN-Folding induced distributional shift can also yield interesting insights on the scaling required at each layer. Note the very different y-axis scales.}
    
\label{fig:mbnet_imgnet_wts}
\end{figure}

Due to space constraints, we ommitted plots of the layerwise weights data from the main paper since the BN-Folded weights were the ones being quantized. However, we can see that even prior to BN-Folding, the weights distributions of MobileNets are poorly behaved. BN-Folding further aggravates this issue. By analyzing the change in range/average precision of weights distributions before/after BN-Folding we can see how certain layers needed greater scaling to properly normalize their representations. This implies a greater misalignment in the learned distributions of that layer. Consequently, we observe even larger BN-Folding-induced distributional shifts in the ImageNet-trained MobileNets models (see Figure~\ref{fig:mbnet_imgnet_wts}). 

We also ommitted the layerwise activation plots as we felt that they followed very similar trends to the BN-Folded weights. Those plots can be seen below in Figures~\ref{fig:cifar_weights},~\ref{fig:cifar_acts}. 

\begin{figure}

    \centering
    \begin{minipage}{0.5\textwidth}
        \centering
        \includegraphics[width=0.95\textwidth]{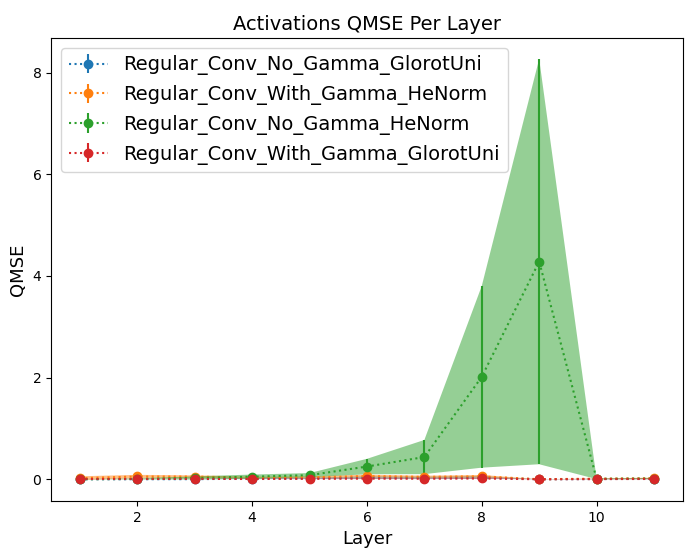} 
    \end{minipage}\hfill
    \begin{minipage}{0.5\textwidth}
        \centering
        \includegraphics[width=0.95\textwidth]{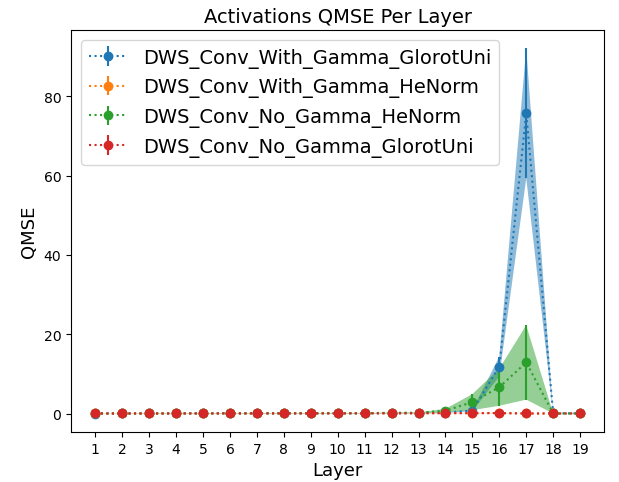}
    \end{minipage}\hfill
    \begin{minipage}{0.5\textwidth}
        \centering
        \includegraphics[width=0.95\textwidth]{MobileNets_Acts_QMSE} 
    \end{minipage}
    \caption{\footnotesize{} Layerwise QMSE for all trained networks. Regular-ConvNets (top), DWS-ConvNets (center), and MobileNets-V1 (bottom). The significant spike for DWS\_Conv\_With\_Gamma\_GlorotUni explain the major degradation in quantized performance. \textbf{Note} the difference in y-axis scales for Regular-ConvNets and DWS-ConvNets. The solid line represents the average values across 5 quantization trials and the shaded region is the standard deviation.}
    
\label{fig:all_qmse}
\end{figure}

Finally, the layerwise QMSE plots in the main paper had some outliers ommitted as those networks' QMSE spiked so high that we could not display the general behaviour of the other networks. The full layerwise QMSE plots with all networks included are in Figure~\ref{fig:all_qmse}. Incidentally, these outlier networks also suffered from the greatest quantization degradation.

\begin{figure*}
    \centering
    \begin{minipage}{0.32\textwidth}
        \centering
        \includegraphics[width=0.98\textwidth]{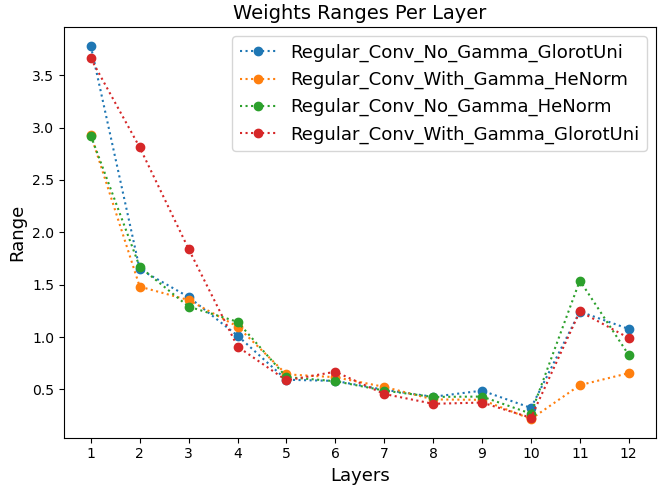} 
    \end{minipage}\hfill
    \begin{minipage}{0.32\textwidth}
        \centering
        \includegraphics[width=0.95\textwidth]{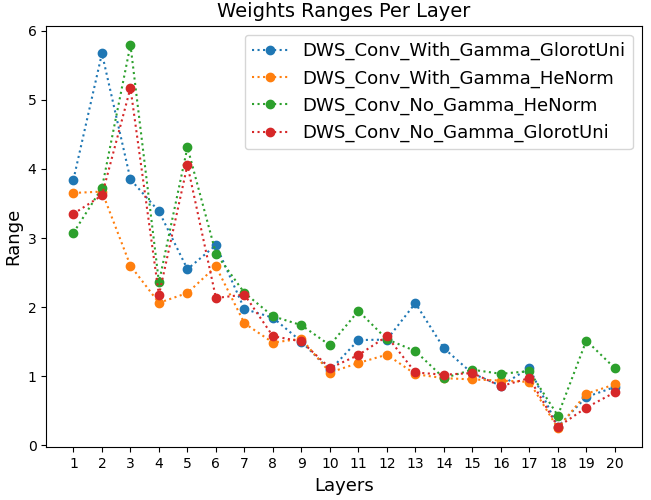}
    \end{minipage}\hfill
    \begin{minipage}{0.31\textwidth}
        \centering
        \includegraphics[width=0.98\textwidth]{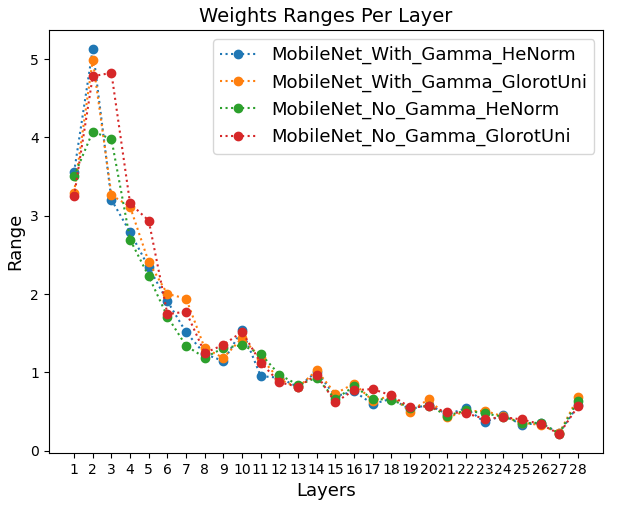} 
    \end{minipage}
    \begin{minipage}{0.31\textwidth}
        \centering
        \includegraphics[width=0.98\textwidth]{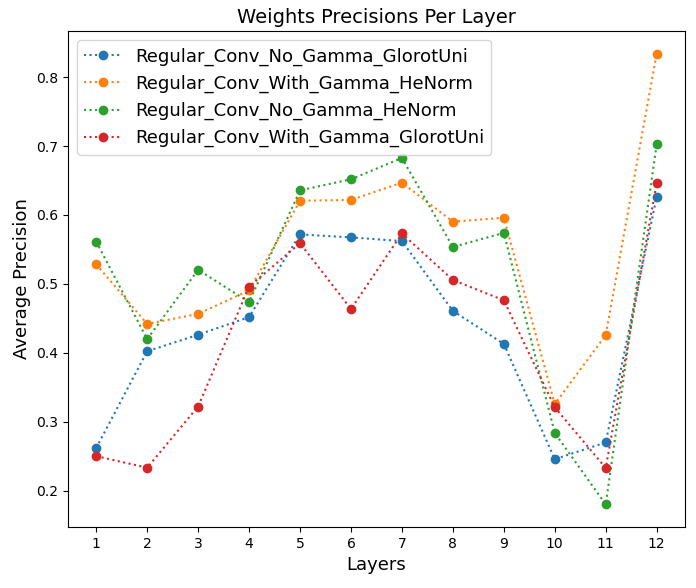} 
    \end{minipage}\hfill
    \begin{minipage}{0.32\textwidth}
        \centering
        \includegraphics[width=0.95\textwidth]{DWS_Conv_Wts_Precisions}
    \end{minipage}\hfill
    \begin{minipage}{0.32\textwidth}
        \centering
        \includegraphics[width=0.95\textwidth]{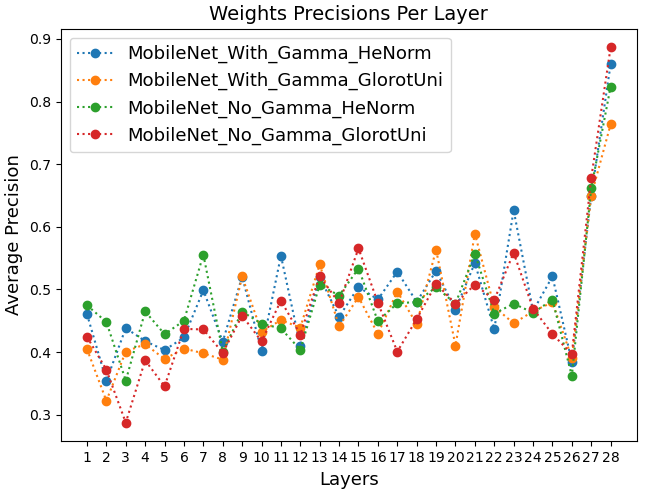} 
    \end{minipage}
    \caption{\footnotesize{} Layerwise weights range (top row) and average precision (bottom row) for all trained networks. Regular CNN (left), DWSCNN (center), and MobileNets-V1 (right).}
    
\label{fig:cifar_weights}
\end{figure*}

\begin{figure*}
    \centering
    \begin{minipage}{0.31\textwidth}
        \centering
        \includegraphics[width=0.98\textwidth]{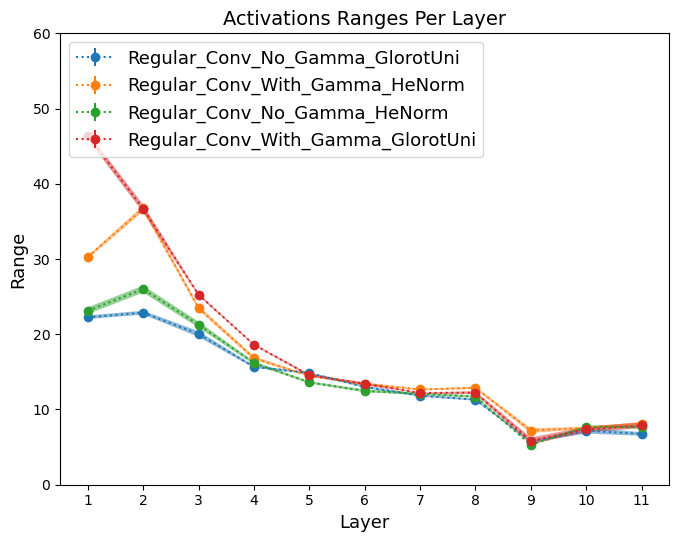} 
    \end{minipage}\hfill
    \begin{minipage}{0.31\textwidth}
        \centering
        \includegraphics[width=0.93\textwidth]{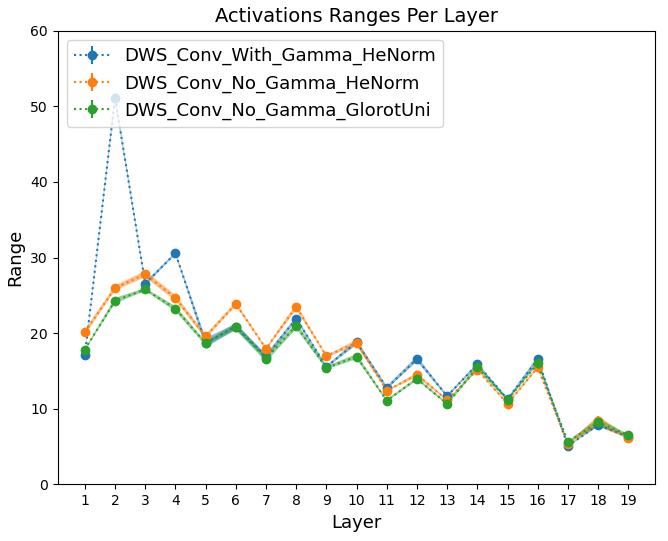}
    \end{minipage}\hfill
    \begin{minipage}{0.31\textwidth}
        \centering
        \includegraphics[width=0.98\textwidth]{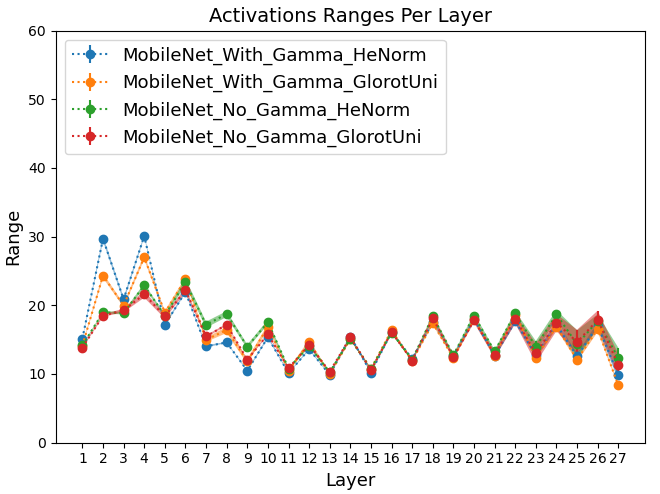} 
    \end{minipage}
    \begin{minipage}{0.31\textwidth}
        \centering
        \includegraphics[width=0.98\textwidth]{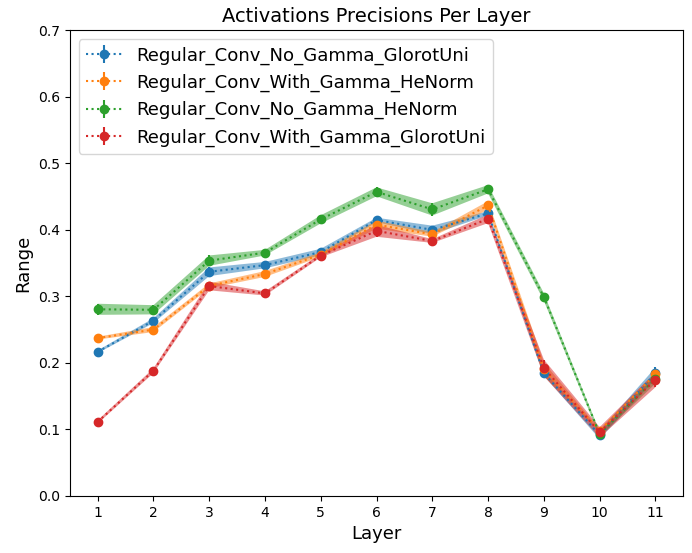} 
    \end{minipage}\hfill
    \begin{minipage}{0.31\textwidth}
        \centering
        \includegraphics[width=0.95\textwidth]{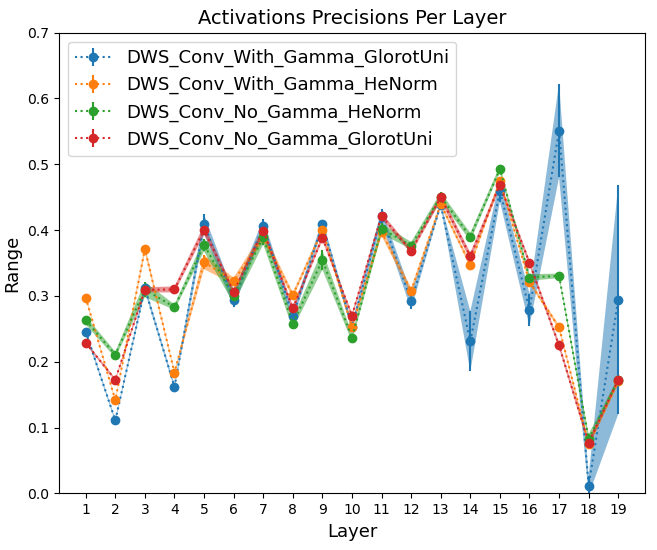}
    \end{minipage}\hfill
    \begin{minipage}{0.31\textwidth}
        \centering
        \includegraphics[width=0.95\textwidth]{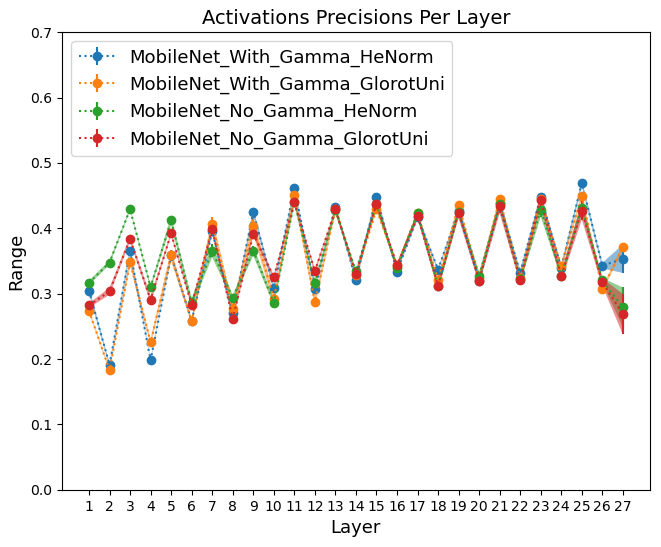} 
    \end{minipage}
    \caption{\footnotesize{} Layerwise activation range (top row) and average precision (bottom row) for all trained networks. Regular CNN (left), DWSCNN (center), and MobileNets-V1 (right). The solid line represents the average values across 5 quantization trials and the shaded region is the standard devation. We can see from the shaded regions that the choice of calibration dataset can lead to non-trivial variations in quantization parameters. \textbf{Note:} DWS-Conv-With-Gamma-GlorotUni has been ommitted from activation range plots due to its extreme outlier activation range of 200 at the first Dense layer (layer 18). This network also had the worst quantization behaviour out of all experiments.}
    
\label{fig:cifar_acts}
\end{figure*}

\subsection{ResNet-34 CIFAR-10 Experiment Details}
\label{sec:resnet34}
In addition to the Regular-Conv and DWS-Conv networks, we also wanted to analyze the multiscale distributional dynamics of a more complex CNN block. Thus, we trained a ResNet-34 based architecture with some modifications for the CIFAR-10 image resolution. Those changes are as follows:
\begin{itemize}
    \item We do not downsample with MaxPooling after the very first conv layer
    \item We do not downsample with stride = 2 at the first residual block with output channels 128
    \item We do not downsample with stride = 2 at the first residual block with output channels 512
    \item Thus, the input tensor shape to Global Average Pooling layer is $4\times4\times1024$ rather than $1\times1\times1024$
\end{itemize}

\begin{figure*}
    \centering
    \begin{minipage}{0.5\textwidth}
        \centering
        \includegraphics[width=0.98\textwidth]{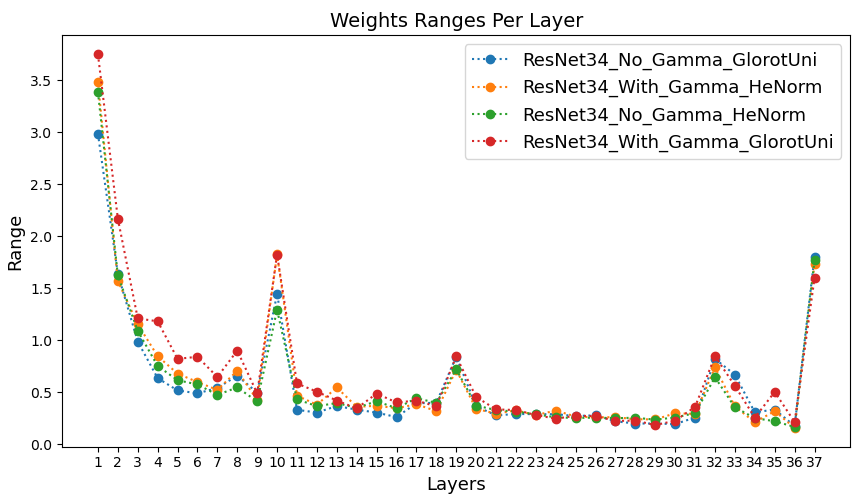} 
    \end{minipage}\hfill
    \begin{minipage}{0.5\textwidth}
        \centering
        \includegraphics[width=0.98\textwidth]{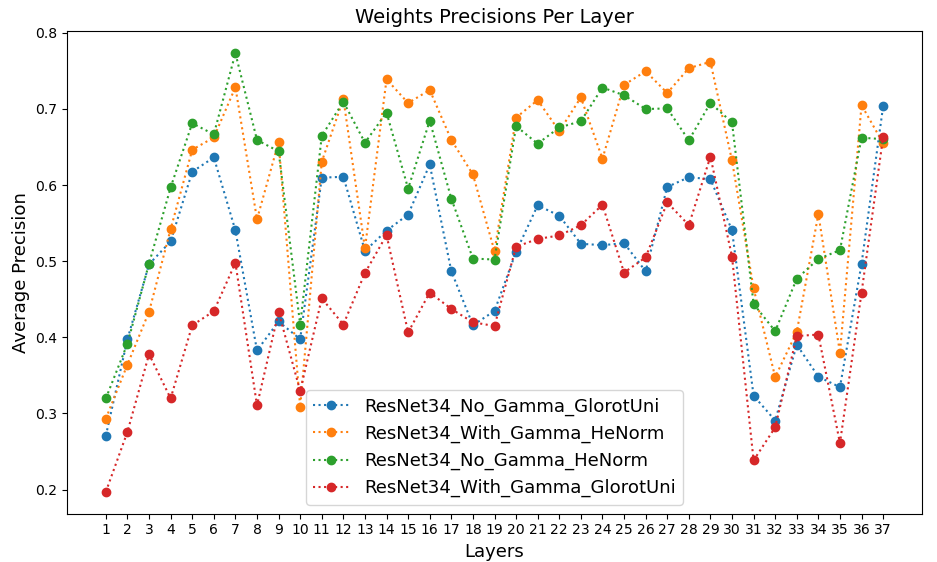} 
    \end{minipage}\hfill

    \caption{\footnotesize{} Layerwise weights range (left) and average precision (right) for ResNet-34 networks.  Note, the spikes observed on the layerwise range plots at layers 10, 19, and 32 correspond to the $1\times1$ linear projection convolution.}
    
\label{fig:resnet_weights}
\end{figure*}

\begin{figure*}

    \centering
    \begin{minipage}{0.5\textwidth}
        \centering
        \includegraphics[width=0.98\textwidth]{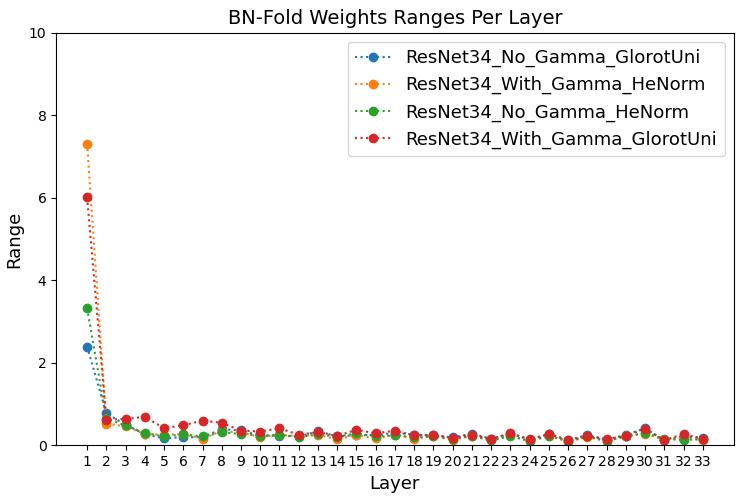} 
    \end{minipage}\hfill
    \begin{minipage}{0.5\textwidth}
        \centering
        \includegraphics[width=0.95\textwidth]{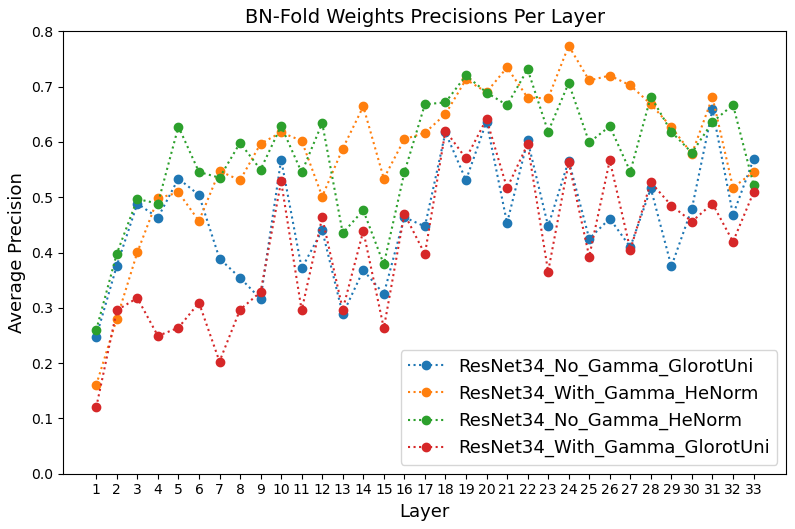}
    \end{minipage}\hfill

    \caption{\footnotesize{} Layerwise BN-folded weights dynamic range (left) and average precision (right) for ResNet-34 networks.}
    
\label{fig:resnet_bn_fold_weights}
\end{figure*}

\begin{figure*}
    \centering
    \begin{minipage}{0.5\textwidth}
        \centering
        \includegraphics[width=0.98\textwidth]{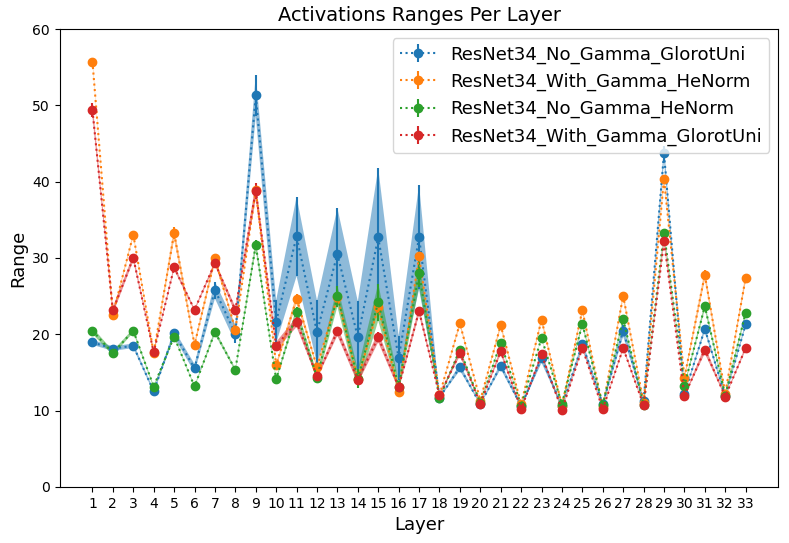} 
    \end{minipage}\hfill
    \begin{minipage}{0.5\textwidth}
        \centering
        \includegraphics[width=0.95\textwidth]{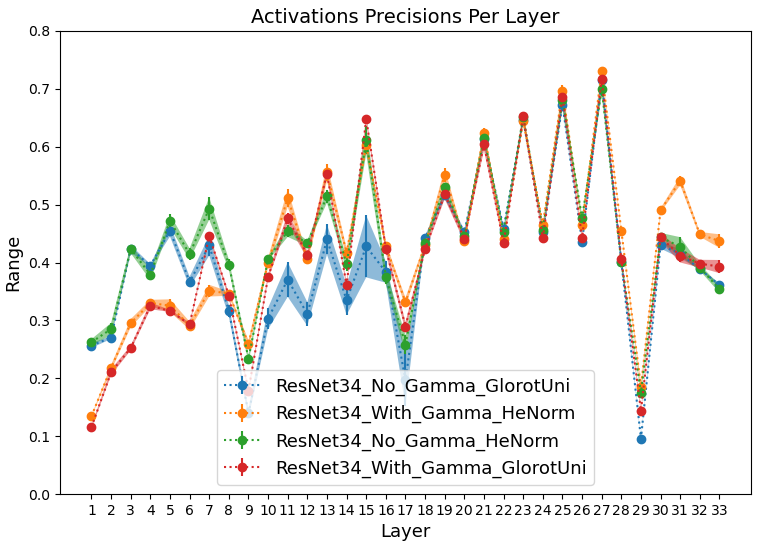} 
    \end{minipage}
    \caption{\footnotesize{} Layerwise activation range (left) and average precision (right) for ResNet-34.}
    
\label{fig:resnet_acts}
\end{figure*}

\begin{figure}

    \centering
    \begin{minipage}{0.5\textwidth}
        \centering
        \includegraphics[width=0.98\textwidth]{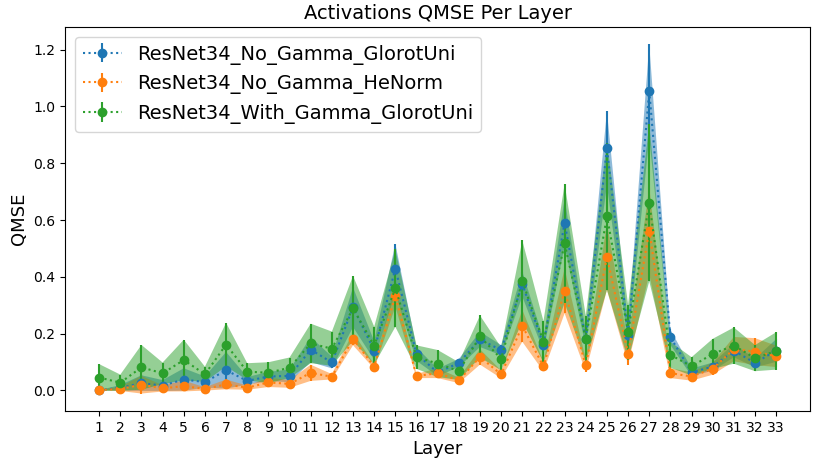}
    \end{minipage}\hfill
    \begin{minipage}{0.5\textwidth}
        \centering
        \includegraphics[width=0.98\textwidth]{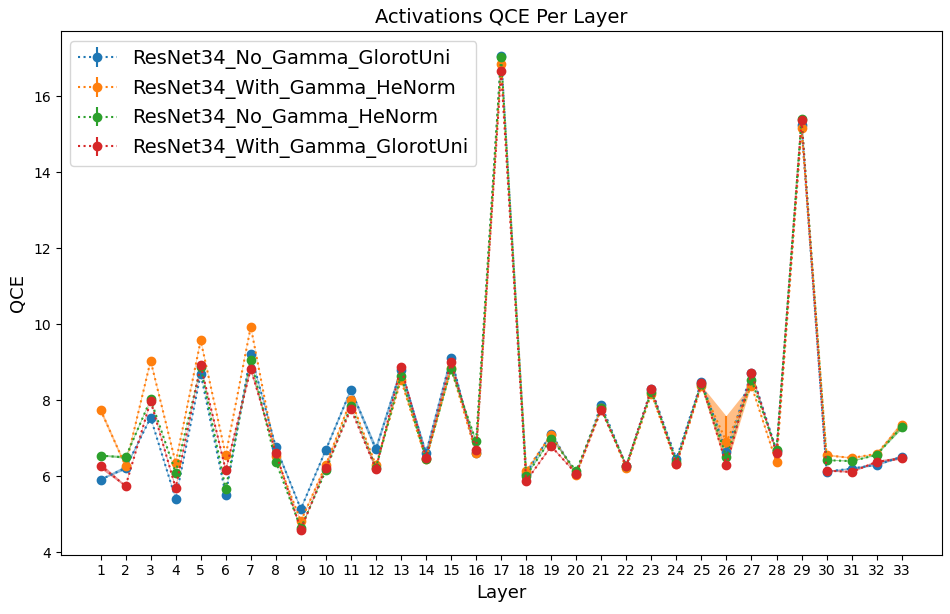}
    \end{minipage}\hfill
    \begin{minipage}{0.5\textwidth}
        \centering
        \includegraphics[width=0.98\textwidth]{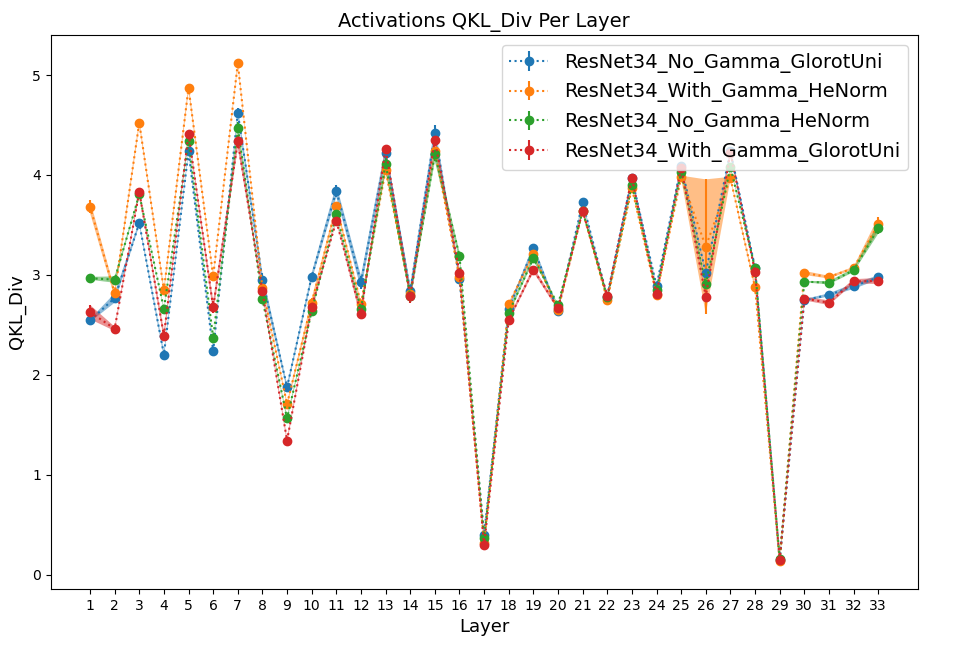}
    \end{minipage}\hfill

    \caption{\footnotesize{} Layerwise QMSE (top) QCE (center) and QKL-Div (bottom) for ResNet-34. The solid line represents the average values across 5 quantization trials and the shaded region is the standard devation. Note, one outlier removed due to large QMSE scale.}
    
\label{fig:resnet_quant}
\end{figure}

The rest of the training details are the same as those mentioned in the Experiments section of the main paper. Full results from all of the CIFAR-10 trained networks are in Table~\ref{full_cifar_quant_results}. Interestingly, the ResNet-34 quantization results are fairly similar to the Regular-Conv Networks. We had originally hypothesized that the skip connection may enable the convolution layers to learn more compact distributions and have less quantization degradation. However, as seen from the quantization results and the plots in Figures~\ref{fig:resnet_weights}--\ref{fig:resnet_quant} this was not necessarily the case. It would appear that while ResNet-34 does indeed learn compact weight distributions (see weights/BN-Fold weights ranges in Figure~\ref{fig:resnet_weights}), the introduction of a skip connection has led to more fluctuations in average precision and possibly offset any potential gains from the smaller dynamic ranges. However, further analysis is required to properly understand this. Overall, it is currently unclear how the introduction of skip connections might affect the quantization dynamics of the system. It would be interesting to see how a ResNet and Regular-ConvNet of equivalent layer-depths behave once trained and quantized. Additionally, comparing bottleneck residual block with inverted bottleneck residual (ie. MobileNet-V2) block should yield further insights on the interplay between reduced range, increased distributional mismatches, and the complex quantization dynamics these systems yield.

\begin{table*}
	\setlength{\tabcolsep}{6pt}
	\centering
	\resizebox{1.01\textwidth}{!}{%
		\csvreader[tabular=|l|c|c|c|c|c|c|,
        table head=\hline \textbf{Network Architecture} & \textbf{FP32 Acc (\%)} & \textbf{QUINT8 Acc (\%)} & \textbf{QMSE} & \textbf{QCE} & \textbf{QKL-Div} & \textbf{Percent Acc Decrease}\\\hline,
        late after line=\\\hline]%
        {resnet34_and_regular_and_mbnet_mean_quant_results.csv}{Network Architecture=\Network, FP32 Acc=\FP, QUINT8 Acc=\QINT, QMSE=\QMSE, QCE=\QCE, QKL-Div=\QKLD, Percent Acc Decrease=\Percent}%
        {\Network & \FP & \QINT & \QMSE & \QCE & \QKLD & \Percent}%
		}
		\caption{Detailed quantization results for each network trained on CIFAR-10. Quantization results are reported as mean and standard deviation across five different quantization trials. The output QMSE, QCE and QKL-Div of depthwise-separable convolution based networks are noticeably higher than regular CNNs in most cases}
	\label{full_cifar_quant_results}
\end{table*}

\subsection{Layerwise Plots From ImageNet Analysis}
\label{sec:imgnet}
\begin{figure}
    \centering
    \begin{minipage}{0.5\textwidth}
        \centering
        \includegraphics[width=0.95\textwidth]{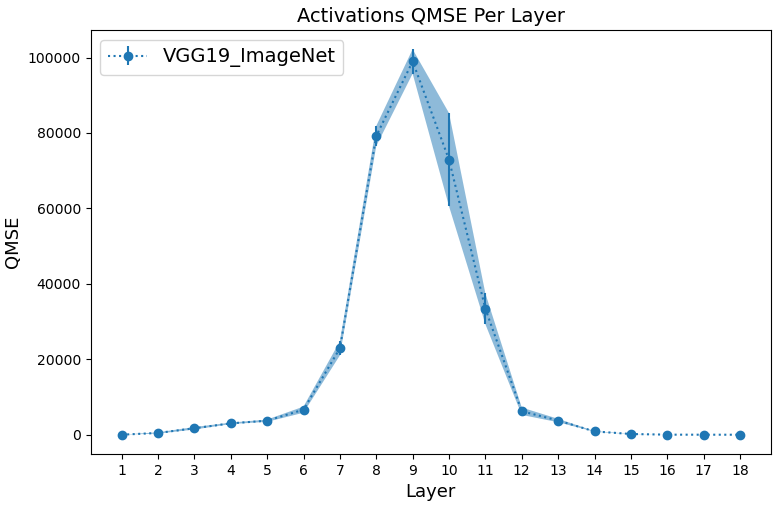} 
    \end{minipage}\hfill
    \begin{minipage}{0.5\textwidth}
        \centering
        \includegraphics[width=0.95\textwidth]{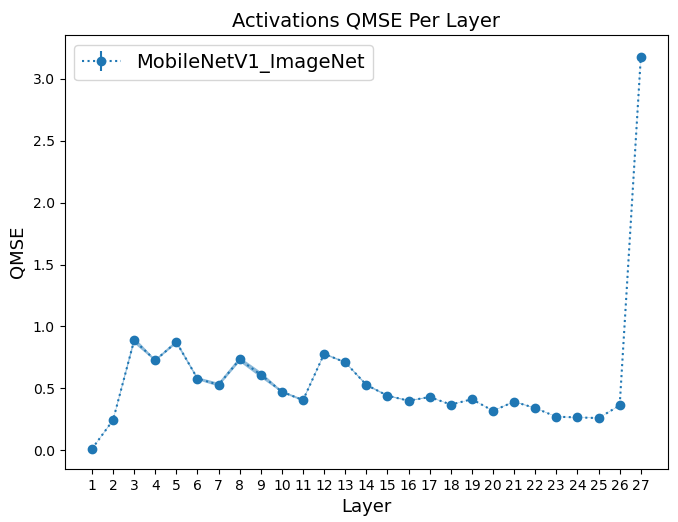}
    \end{minipage}\hfill
    \caption{\footnotesize{} Layerwise QMSE for VGG-19 (top) and MobileNet-V1 (bottom) trained on ImageNet. Note the difference in y-axis scales.}
    
\label{fig:imgnet_qmse}
\end{figure}

\begin{figure}

    \centering
    \begin{minipage}{0.5\textwidth}
        \centering
        \includegraphics[width=0.98\textwidth]{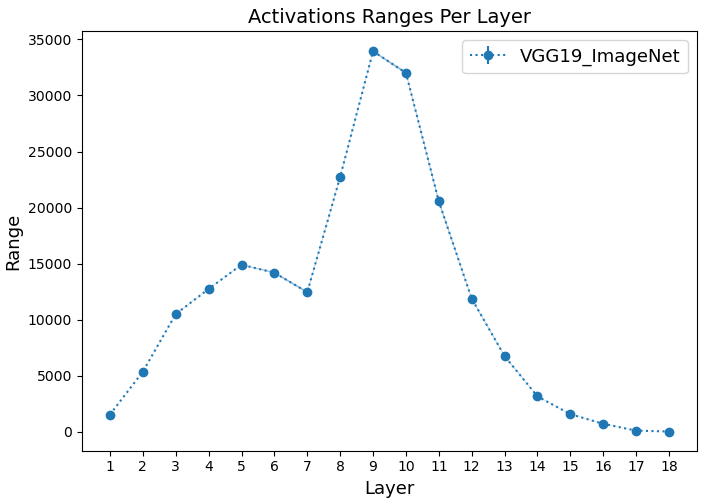} 
    \end{minipage}\hfill
    \begin{minipage}{0.5\textwidth}
        \centering
        \includegraphics[width=0.98\textwidth]{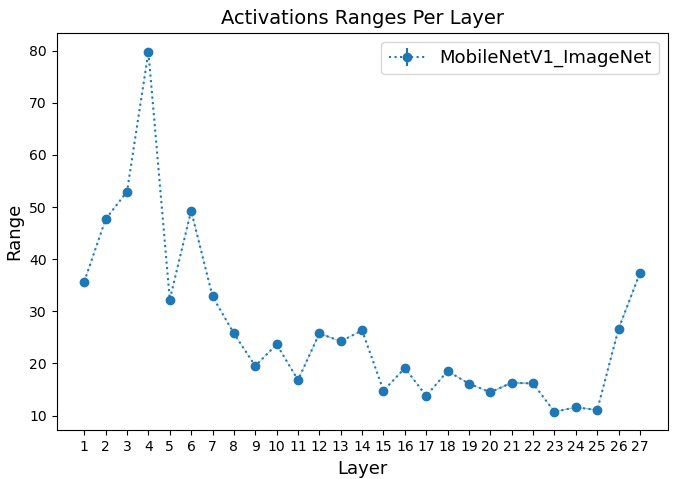}
    \end{minipage}\hfill
    \begin{minipage}{0.5\textwidth}
        \centering
        \includegraphics[width=0.98\textwidth]{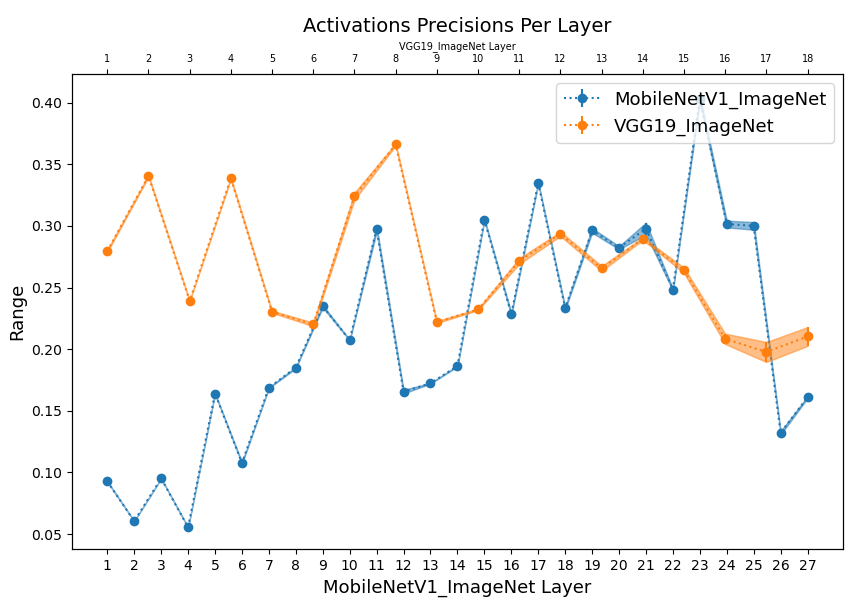} 
    \end{minipage}
    \caption{\footnotesize{} Layerwise activation ranges for VGG-19 (top) and MobileNet-V1 (center) trained on ImageNet. Layerwise activation precisions for VGG-19 vs. MobileNet-V1 trained on ImageNet (bottom).}
    
\label{fig:imgnet_act_stats}
\end{figure}

Here, we've included all of the ommitted layerwise plots from the fine-grained, multiscale analysis of the ImageNet trained networks. As mentioned in the main paper, it was hard to compare the layerwise QMSE and Activations data. The differences in preprocessing for the two networks led to widely different scales/ranges. Worth noting is the overall trends in these distributions. QMSE increases and then decreases in VGG-19 while for MobileNet-V1 it stays generally high (compared to our CIFAR-10 networks which were preprocessed the same way. Eg. Normalized to the range [-1, 1]) and spikes near the end (see Figure~\ref{fig:imgnet_qmse}). Furthermore, we still observe similar trends in fluctuating ranges/average precisions from the layerwise activations plots in Figure~\ref{fig:imgnet_act_stats}. Thus, further supporting our hypothesis that depthwise-separable convolutional networks tend towards learning mismatched distributions, regardless of training data. 

Worth noting is that the diversity of a large-scale dataset like ImageNet also seems to have increased mismatch of distributions for VGG-19 (see in Figure~\ref{fig:imgnet_act_stats} as well as ImageNet results in the main paper). Though the distributions of VGG-19 are better behaved in general, we see that regular-conv networks might also benefit from better-aligned distributional dynamics and the quantized accuracy drop could be related to the drop in precision plus spike in range we observe for both the weights and activations near the ``middle" of VGG-19.

\end{document}